%% file: main.tex
\documentclass[10pt,twocolumn,letterpaper]{article}
\usepackage{ifthen}
\newboolean{public_version}
\setboolean{public_version}{false} 
\usepackage[pagenumbers]{cvpr}

\input{preamble}
\definecolor{cvprblue}{rgb}{0.21,0.49,0.74}
\usepackage[pagebackref,breaklinks,colorlinks,allcolors=cvprblue]{hyperref}
\newcommand{\PaperTitle}{Unlocking ImageNet’s Multi-Object Nature: Automated Large-Scale Multilabel Annotation}

\title{\PaperTitle}

\author{Junyu Chen$^{1}$
\qquad Md Yousuf Harun$^{2}$ \qquad Christopher Kanan$^{1}$\\
$^1$University of Rochester \qquad $^2$Rochester Institute of Technology \\
}

\begin{document}
\maketitle

\begin{abstract}
The original ImageNet benchmark enforces a single-label assumption, despite many images depicting multiple objects. This leads to label noise and limits the richness of the learning signal. Multi-label annotations more accurately reflect real-world visual scenes, where multiple objects co-occur and contribute to semantic understanding—enabling models to learn richer and more robust representations. While prior efforts (e.g., ReaL~\cite{real}, ImageNetv2~\cite{shankar2020evaluating}) have improved the validation set, there has not yet been a scalable, high-quality multi-label annotation for the training set. To this end, we present an automated pipeline to convert the ImageNet training set into a multi-label dataset—without human annotations. Using self-supervised Vision Transformers, we perform unsupervised object discovery, select regions aligned with original labels to train a lightweight classifier, and apply it to all regions to generate coherent multi-label annotations across the dataset.
Our labels show strong alignment with human judgment in qualitative evaluations and consistently improve performance across quantitative benchmarks. Compared to traditional single-label scheme, models trained with our multi-label supervision achieve consistently better in-domain accuracy across architectures (up to $+2.0$ top-1 accuracy on ReaL and $+1.5$ on ImageNet-V2) and exhibit stronger transferability to downstream tasks (up to $+4.2$ and $+2.3$ mAP on COCO and VOC, respectively). These results underscore the importance of accurate multi-label annotations for enhancing both classification performance and representation learning. Project code and the generated multi-label annotations are available at \url{https://github.com/jchen175/MultiLabel-ImageNet}.
\end{abstract}

\input{figures/relabel_example}

\section{Introduction}
The ImageNet-1K dataset~\cite{imagenet} has long served as a cornerstone for computer vision. Its impact spans not only vision models~\cite{he2016deep,oquab2023dinov2,simeoni2025dinov3} but also multimodal systems that use it for visual pretraining and evaluation~\cite{radford2021learning,li2022blip,alayrac2022flamingo}. A known limitation, however, is its single-label assumption—each image is annotated with only one category, even though many depict multiple objects or concepts~\cite{real,shankar2020evaluating,anzaku2024re}. This single-label assumption often misrepresents the image content and introduces label noise. Prior analyses have shown that a significant portion of ImageNet images fall into these problematic cases: images with multiple valid labels, synonymous or hierarchically overlapping labels, or truly incorrect labels~\cite{ridnik2021imagenet,vasudevan2022does,recht2019imagenet,relabel}. In fact, nearly $15\%$ of the images were found to contain at least two relevant categories when re-examined by human annotators~\cite{real}. Such findings underscore that ImageNet’s single-label annotations are often misaligned with the dataset’s inherently multi-label nature.
These labeling issues have serious implications for both model training and evaluation. During training, incomplete or wrong label set yields noisy or incorrect supervision that can hinder learning~\cite{relabel}. During evaluation, models are penalized for predicting secondary objects present in the image, since only one ``ground-truth'' label is provided~\cite{real}. This not only unfairly penalizes accurate predictions of additional objects but also complicates model benchmarking. For example, recent studies confirmed that much of the perceived drop in ImageNet-V2~\cite{recht2019imagenet} accuracy is explained by its higher fraction of multi-object images, rather than fundamental model degradation~\cite{anzaku2024re,anzaku2023leveraging}. 

Recent work has addressed this gap for evaluation: ImageNet-ReaL~\cite{real} and Multilabelfy~\cite{anzaku2023leveraging} provide human-verified multi-label annotations for the validation set. These enable more accurate benchmarking and reveal significant label omissions in the original dataset. However, little progress has been made toward relabeling the training set—largely due to the prohibitive cost of manually re-annotating 1.28M images~\cite{shankar2020evaluating}.
One notable approach is ReLabel~\cite{relabel}, which sidesteps manual labeling by using a strong classifier to produce a patch-wise label map for each image. At training time, random crops are supervised via pooled labels from this map, providing soft, localized supervision that improves performance. However, ReLabel’s supervision is limited to single soft-labels per crop: it does not provide an explicit set of all object classes per image, nor instance-level separation. \emph{Thus, despite prior work, there is still no publicly available ImageNet-1K training set with complete multi-label annotations.} Such a dataset would allow models to learn from all objects within each image—reflecting the true complexity of real-world scenes.

In this work, we aim to bridge this gap by producing a fully multi-labeled version of the ImageNet-1K training set.  Rather than relying on global classifiers—which is prone to overfit—we explicitly localize object instances and assign labels at the region level. Leveraging recent advances in self-supervised learning (SSL) pretraining~\cite{simeoni2025dinov3} and unsupervised object detection~\cite{wang2023cut}, we identify candidate regions in each image, covering salient regions that likely correspond to objects, and treat any other masks in the image at this stage as unlabeled instances. Using the confirmed primary regions and their known class labels, we then train a region-based classifier to recognize object crops in a category-specific manner. In effect, the model learns to predict accurate ImageNet class given localized patch features. This step is crucial to prevent the classifier from shortcut learning the original single label from background or contextual cues. Finally, we deploy this refined classifier on all discovered object proposals in each image. Consequently, we generate comprehensive multi-label annotations across the entire ImageNet-1K training set with object-level grounding (see Fig.~\ref{fig:example}).

We evaluate our annotations through extensive experimentation. Qualitatively, our labels consistently correspond well with presented objects and improve semantic alignment with the image content. Quantitatively, models trained on our multi-label annotations show improved validation performance (up to $+2.0$ top-1 accuracy on ReaL~\cite{real} and $+1.5$ ImageNet-V2~\cite{recht2019imagenet}). Moreover, in transfer learning to downstream multi-label tasks, models pre-trained on our multi-label dataset consistently outperform single-label baselines (up to
$+4.2$ and $+2.3$ mAP on COCO~\cite{lin2014microsoft} and VOC~\cite{everingham2010pascal}, respectively), highlighting the transferability of features learned from richer supervision.

\noindent \textbf{In summary, our contributions are threefold:}
\begin{itemize}
    \item \textbf{Automated large-scale multi-label annotation.} 
    We introduce a fully automated pipeline that generates explicit multi-label annotations for all $1.28$M ImageNet-1K training images—without human labeling. To our knowledge, this is the first work to produce dense multi-label annotations at this scale. The pipeline is general and can convert other single-label datasets into multi-label form.

    \item \textbf{Improved label quality and instance attribution.} 
    Our annotations recover missing classes overlooked by prior efforts such as ReaL and associate each label with a localized object region. When combined with ReaL, our labels reduce false negatives and correct inconsistencies, offering a scalable and interpretable relabeling strategy.
    
    \item \textbf{Better supervision and transferability.} Models trained with our multi-label annotations achieve consistent gains on both in-distribution and  downstream multi-label benchmarks, surpassing single-label and single-positive learning baselines. Improvements hold across diverse architectures and scales—from ResNet-50 to ViT-Large—demonstrating the robustness of our supervision.
\end{itemize}

\section{Background}
\subsection{Re-labeling ImageNet and Dataset Quality}
The shortcomings of ImageNet’s single-label annotations have been widely recognized. Early analyses~\cite{stock2018convnets,recht2019imagenet} uncovered label noise and generalization gaps, with ImageNet-V2 exposing an $11$–$14\%$ accuracy drop. Subsequent studies~\cite{shankar2020evaluating} showed that many apparent errors stemmed from valid secondary objects missing in the ground truth. ReaL~\cite{real} addressed this by providing multi-label annotations for the validation set, enabling more accurate evaluation. Performance improved significantly under this protocol, underscoring the importance of label completeness. Multilabelfy~\cite{anzaku2023leveraging} confirmed that nearly half of ImageNet-V2 images have multiple valid labels, and that multi-label evaluation better reflects true model performance.
Efforts to improve the training set have been more limited. Human re-annotation at ImageNet’s scale ($1.2$M images) is prohibitively expensive~\cite{shankar2020evaluating}. Existing work has largely focused on automated pipelines. ImageNet-Segments~\cite{gao2022large} adds pixel-level masks for a subset of training images but only labels one object per image. MIIL~\cite{ridnik2021imagenet} derives semantic multi-labels from ImageNet-21K via WordNet hierarchies, but lacks spatial grounding. ReLabel~\cite{relabel}, the most relevant to our work, uses a pretrained classifier to generate soft spatial labels, supervising random crops via pooled logits. While effective, it still assumes a single soft label per region and does not yield explicit image-level multi-labels.
In contrast, our method produces discrete multi-label annotations grounded to object proposals for every training image (Fig.~\ref{fig:example}). This leads to more interpretable and complete labels, enabling stronger supervision and improved transfer learning.

\input{figures/diagram}

\subsection{Multi-Label Learning from Single Labels}
Our work also connects to weakly supervised multi-label learning, where the training data contain incomplete or single-label annotations. ImageNet represents an extreme case: each image has only one label despite often depicting multiple objects. Recent methods attempt to recover missing labels from such data. Spatial Consistency Loss (SCL)~\cite{verelst2023spatial} maintains a moving average of class activation maps as evolving pseudo-labels, enforcing consistency under augmentations to uncover additional objects. Large Loss (LL)~\cite{kim2022large} analyzes partial-label training dynamics and identifies a memorization effect—where models first learn true positives, then overfit by treating unannotated labels as negatives. They mitigate this by down-weighting high-loss dimensions, which likely correspond to missing objects. While these approaches can partially recover multi-label signals, our goal differs: we explicitly relabel the entire ImageNet-1K training set with region-grounded multi-label annotations. In our experiments, we compare against these methods and show that explicit supervision yields stronger performance.

\subsection{Unsupervised Object Discovery}
Unsupervised object discovery seeks to localize objects without human annotations, and recent self-supervised learning advances have greatly improved this task. TokenCut~\cite{wang2023tokencut} pioneered this direction by leveraging DINO~\cite{dino} ViT features to construct a patch similarity graph and applying Normalized Cut to segment the most salient object—achieving strong results without training, but limited to one object per image. CutLER~\cite{wang2023cut} extends this to multi-object discovery via MaskCut, which iteratively masks out detected regions and reapplies TokenCut to uncover additional objects. These coarse masks are then refined through self-training. We adopt MaskCut in our pipeline to generate candidate object masks on ImageNet, forming the basis for relabeling. Compared to general segmentation tools like Segment Anything (SAMv2)~\cite{ravi2024sam}, MaskCut provides more consistent object-level proposals (Supplementary Material~\ref{app:sam}; abbreviated as Supp. throughout).

\section{Relabeling ImageNet}
Our pipeline is illustrated in Fig.~\ref{fig:diagram} (a,b) which consists of three stages: (1) unsupervised object discovery to generate proposals~\footnote{We sometimes refer to object proposals as masks}, (2) training a classification head on selected regions, and (3) inferring multi-label annotations by aggregating per-mask predictions. Below we outline each stage.
\subsection{Unsupervised Object Mask Discovery}
To localize candidate object regions without using labels, in each image $I \in \mathbb{R}^{h \times w}$, we adopt MaskCut~\cite{wang2023cut}
to ViT patch embeddings extracted from the penultimate layer.
After each iteration, MaskCut produces a binary mask $P^\prime_i \in \{0,1\}^{h' \times w'}$ at the resolution of the ViT feature map, indicating a candidate object region. We then apply refinement steps from CutLER~\cite{wang2023cut} (including CRF~\cite{sutton2012introduction} post-processing) which upsample the mask to the original image resolution $P_i \in \{0,1\}^{h \times w}$ (detailed in Supp.~\ref{app:mcut_detail}). Given a self-supervised ViT encoder $\mathcal{F}$ (e.g., DINOv3~\cite{simeoni2025dinov3}), we extract up to $N$ proposals per image:

\vspace{-1mm}
\[
{P_1, P_2, \ldots, P_N} = \texttt{MaskCut}(\mathcal{F}, I),
\vspace{-1mm}
\]

While general-purpose models like SAM~\cite{ravi2024sam} offer flexible mask generation, we found MaskCut more suitable for consistent object-level proposals due to its stability and scalability (see Supp.~\ref{app:sam} for comparisons to SAM).

\noindent \textbf{Hyperparameter Tuning.}
We sweep key parameters—including the affinity threshold, number of proposals $N$, and ViT backbone—and evaluate object recall on a small validation set. The top-performing configurations are manually inspected, and the best $4$ are selected for use in our final pipeline. Full tuning protocol and results are provided in Supp.~\ref{app:mcut_sweep}.

\subsection{Localized Labeler Training} 
\label{sec:labeler_training}

Given object proposals, our goal is to assign class labels to these regions and train a classifier capable of predicting multiple objects per image. Naively label every proposal with the image-level ground-truth $y$ leads to severe overfitting, causing the classifier to predict $y$ even for background or irrelevant regions (e.g., EVA02~\cite{fang2024eva} in Fig.~\ref{fig:diagram}(c)).

To obtain reliable supervision, we use ReLabel~\cite{relabel}, which provides a $[15 \times 15 \times 5]$ grid of top-5 class indices and corresponding logits per image. Following their implementation, we convert this sparse representation into a dense $15 \times 15 \times 1000$ logit tensor $Z$ by inserting each top-5 logit at its respective class index, leaving all other entries zero. This produces a per-location class logit map, which we then bilinearly upsample to the full image resolution $(h \times w)$ to obtain pixel-wise logits $Z \in \mathbb{R}^{h \times w \times 1000}$.

Given a proposal mask $P \in \{0,1\}^{h \times w}$ and the pixel-level logits $Z_{pq}[c]$, we compute its logit vector $v_P \in \mathbb{R}^{1000}$ by masking the logit map and averaging over the foreground pixels:
\vspace{-2mm}
\[
v_P[c] = 
\frac{1}{\sum_{p,q} P_{pq}}
\sum_{p,q} \left(P \odot Z[c]\right)_{pq},
\vspace{-1mm}
\]
where $\odot$ is Hadamard product
and $Z[c]$ denotes the logit map for class $c$.
After applying a softmax, we obtain the class probability distribution: $s_P = \texttt{softmax}(v_P)$.
We retain only proposals whose confidence on the image's original label $y$ exceeds a threshold, i.e., $s_P(y) > \tau_{\text{sel}}$, thereby filtering out unrelated 
regions (see Fig.~\ref{fig:diagram}(c)).

Next, we train a lightweight classification head $h(\cdot)$---a 2-layer MLP with hidden dimension 1024---on top of a frozen DINOv3 ViT-L/16 backbone $\mathcal{F}$.
For each retained proposal $P$ in image $I$, 
we extract patch features $F = \mathcal{F}(I) \in \mathbb{R}^{32 \times 32 \times 1024}$,
and project its mask to patch resolution 
($M \in \{0,1\}^{32 \times 32}$). Treating $F_{ij} \in \mathbb{R}^{1024}$ as the embedding at patch $(i,j)$, we compute the pooled feature
\vspace{-2mm}
\[
z_P =
\frac{1}{\sum_{i,j} M_{ij}}
\sum_{i,j} (M \odot F)_{ij}
\quad \in \mathbb{R}^{1024},
\vspace{-2mm}
\]
i.e., a masked average over the foreground patches by broadcasting $M$ along the channel dimension.
The classification head produces logits $h(z_P) \in \mathbb{R}^{1000}$, trained 
with cross-entropy loss using the original image label $y$. This yields a region-level classifier that generalizes beyond the primary label and enables accurate multi-label prediction (see Fig.~\ref{fig:diagram}(d)). Additional details are in Supp.~\ref{app:labeler_training_details}.

\input{figures/qualitative}
\subsection{Multi-Label Inference via Mask Aggregation}
At inference time, we apply the trained labeler to each object proposal from MaskCut to generate multi-label predictions. For each mask $P_i$, we compute the pooled feature $z_{P_i}$ and obtain a softmax distribution over 1000 classes via the classification head $h$. We then extract the top-1 class prediction $\hat{c}_i = \arg\max h(z{P_i})$ with confidence $\alpha_i$. To form image-level labels, we aggregate all top-1 predictions across proposals—retaining only unique classes and keeping the highest confidence when duplicates occur. This produces a set of spatially grounded labels per image. We optionally filter low-confidence predictions and report the resulting label distribution. As summarized in Supp.~\ref{app:train_set_relabel_statistics} Table~\ref{tab:train_statistics}, over 20\% of training images contain confident multiple labels, highlighting the prevalence of multi-object scenes and the importance of moving beyond single-label supervision.

\section{Human-Verified Multi-Label Comparison}
We evaluate our relabeling pipeline against ReaL~\cite{real}, a human-verified multi-label benchmark for the ImageNet validation set. ReaL aggregates predictions from $19$ trained models, selects $6$ high-recall models to propose candidate labels, and uses human annotators to validate them. Notably, images for which all models agreed on the original label ($25,111$ in total) were not re-annotated. The final ReaL dataset contains $57,553$ verified labels over $46,837$ images, with $3,163$ images left unlabeled. To assess alignment, we binarize our model’s softmax outputs with a threshold of $0.5$ and categorize each image based on the overlap with ReaL into five groups: (1) no labels from ReaL; (2) exact match; (3) our predictions are a superset; (4) ReaL is a superset; and (5) partial overlap. We sampled $50$ images from each group ($250$ total) for human evaluation. 
A detailed qualitative analysis is provided in Supp.~\ref{app:qualitative}, with a breakdown of agreement between our relabeling
and ReaL in Table~\ref{tab:qualitative}

From this evaluation, we demonstrate that our relabeling pipeline effectively improves label coverage and grounding. Among the $3,163$ validation images with no ReaL labels (see Fig.~\ref{fig:qualitative}.(a)), we estimate $64.0\%$ contain valid objects, and our method correctly recovers over $90\%$ of them. For the $12.3\%$ of images where ReaL missed one or more valid labels, our pipeline correctly added them in $84\%$ of cases. These estimates highlight a key limitation of ReaL’s high-precision approach—particularly its omission of human review when models agree—and emphasize the value of explicit multi-label annotations. Furthermore, for images where our labels exactly matched ReaL, $94\%$ of our predicted regions accurately localized the target object (see Fig.~\ref{fig:qualitative}.(a)), confirming strong spatial grounding. Finally, in $5.8\%$ of cases where ReaL included potentially extraneous or debatable labels, our model produced more conservative and accurate predictions.

\noindent \textbf{Remark.} 
Our method assumes one label per region, which holds for datasets like COCO~\cite{lin2014microsoft} or VOC~\cite{everingham2010pascal}, but can fail under ImageNet’s taxonomy(Fig.~\ref{fig:qualitative}.(b))—for example, with synonyms (e.g., sunglass vs. sunglasses), part–whole pairs, or hierarchical classes. We identify 26 such ambiguous class pairs and propose two fixes using co-occurrence priors from ReaL (Supp.~\ref{app:fix_ambiguous_class}; Table~\ref{tab:ambiguous_pairs}, Table~\ref{tab:fix_ambiguous}). While effective, these rely on ReaL statistics and may bias evaluation, so we exclude them from main results.

\input{tables/main_exp}
\input{tables/main_exp_subgroup_real}

\section{Quantitative Experiments}
We quantitatively evaluate the efficacy of our multi-label ImageNet relabeling. First, we compare strategies for converting patch-level outputs into image-level labels (hard vs. soft), and investigate whether including the original image-level label improves performance (Sec.~\ref{sec:sweep_label}). We then compare models trained with our multi-label against those trained with single-label and recent single-positive learning baselines (Sec.~\ref{sec:main_results},~\ref{sec:arch_size_transfer}). Finally, we assess the transferability of our labels by fine-tuning pretrained models on standard multi-label benchmarks (Sec.~\ref{sec:arch_size_transfer}). 

\subsection{Datasets and Metrics}
\noindent \textbf{ImageNet and variants.} ImageNet-1K (IN;~\cite{imagenet}) contains $1,000$ classes with $1.2$M training images and $50$K validation images, each annotated with a single label. ImageNet-Segmentation (IN-Seg;~\cite{gao2022large}) extends this by providing pixel-wise annotations for $40$K validation images across $919$ object categories; we use its $11$K public validation set for evaluation. We also evaluate on ReaL~\cite{real}, a human-verified multilabel version of the ImageNet validation set ($47$K images), and on ImageNet-V2 (INv2)~\cite{recht2019imagenet}, a $10$K-image test set for generalization. Additionally, we use INv2-Multilabelfy (INv2-ML~\cite{anzaku2023leveraging}), which adds human-verified multilabel annotations to INv2, revealing that $47.9\%$ of its images contain multiple valid labels.

\noindent \textbf{Multi‑label benchmarks.} Pascal VOC 2007 (VOC,~\cite{everingham2010pascal}) has $9,963$ images across $20$ object classes; we train on the $5,011$-images train/val split and evaluate on the $4,952$-images test set. MS COCO 2017 (COCO,~\cite{lin2014microsoft}) includes $118$K training and $5$K validation images across $80$ object classes; we use the standard train/val split.

\noindent \textbf{Evaluation metrics.} On ImageNet and INv2, we report top-1 accuracy under both single-label (standard) and multi-label (any correct label counts) criteria~\cite{real,anzaku2023leveraging,shankar2020evaluating}. The top-1 prediction is taken as the class with the highest softmax score (i.e., the argmax of probabilities over the class dimension). For IN-Seg, ReaL, and INv2-ML, we also report mean Average Precision (mAP). For VOC and COCO, we follow standard practice and report mAP across all classes.

\subsection{Comparing Label Aggregation Strategies}
\label{sec:sweep_label}
Our multi-label annotations assign each object proposal $M_i$ a soft class probability vector $\mathbf{p}_{M_i} \in [0,1]^{K}$ where $K=1000$ for 1000 ImageNet classes. We compare two aggregation strategies to construct image-level training labels from these mask-level scores:
 
\noindent \textbf{Local-Hard.}
We apply a threshold $\tau$ to the per-mask probabilities and include any class whose maximum score across all masks exceeds $\tau$. This yields a multi-hot label vector $\hat{y} \in \{0,1\}^K$ indicating the presence of $K$ classes in the image. \textit{Example:} Suppose an image contains two proposals with class probabilities $M_1$: cat = 0.85, $M_2$: dog = 0.72. If $\tau = 0.8$, only \texttt{cat} is included in $\hat{y}$.
 
\noindent \textbf{Local-Soft.}
We aggregate scores by taking the element-wise maximum across masks: $\hat{y}[c] = \max_i \mathbf{p}_{M_i}[c]$, resulting in a soft label vector $\hat{y} \in [0,1]^K$ . This way we preserve relative confidences without thresholding.
\textit{Example:} From the same mask predictions above, the soft label vector would have $\hat{y}[\text{cat}] = 0.85$ and $\hat{y}[\text{dog}] = 0.72$.

\noindent \textbf{Adding Global Signal.} Since localized labels may miss global cues, we explore incorporating an additional global signal $y^{\text{global}}$—either the original single-label ImageNet annotation (\textit{Original}) or the prediction from our classification head applied to the globally pooled encoder features (\textit{Pred}). Final labels are computed as
$\tilde{y}^{\text{final}}[c] = \max\bigl(\tilde{y}^{\text{local}}[c], y^{\text{global}}[c]\bigr),$
where $y^{\text{global}}[c] = 1$ if class $c$ is present in the global label, or is set to the corresponding classifier probability otherwise.

\noindent \textbf{Results.}  
We evaluate each label aggregation strategy by training a ResNet-50~\cite{he2016deep} for 100 epochs and measuring performance on ImageNet evaluation datasets. For \textit{Local-Hard} labels, we sweep the threshold $\tau$ to select the optimal value. Full training details are provided in Supp.~\ref{app:sweep_label_setup}. Results are summarized in Table~\ref{tab:sweep_label_setup} in Supp. We find that \textit{Local-Soft} outperforms \textit{Local-Hard}, and combining global signal improves accuracy further. Specifically, using the original ImageNet label as the global signal yields slightly better results than using our classifier’s global prediction ($+0.2$ accuracy on IN-Val and ReaL), likely because our classifier is better suited for localized predictions. 
Across all comparisons, our multi-label annotation consistently outperforms the original single-label supervision, yielding average gains of $+0.96$ accuracy and $+1.16$ mAP across validation sets.
Based on these results, we adopt \textit{Local-Soft + Original} as the default label setup in all subsequent experiments.

\input{tables/cross_arch_size_224}

\subsection{ImageNet Classification}
\label{sec:main_results}
We evaluate the effectiveness of our relabeled ImageNet by training a ResNet‑50~\cite{he2016deep} from scratch using BCE loss and comparing various training strategies:
\begin{itemize}
    \item \textbf{Baseline}: Standard single-label training with one-hot encoded targets.
    \item \textbf{Large Loss (LL)}~\cite{kim2022large}: LL is a single-positive multi-label method that treats all unobserved classes as negatives, but down-weights dimensions with large training losses to mitigate false-negative memorization. It is fine-tuned from a single-label pretrained model.
    \item \textbf{Spatial Consistency Loss (SCL)}~\cite{verelst2023spatial}: SCL adds a temporal consistency loss encouraging class heatmaps to remain stable across random crops and training epochs. It is also fine-tuned from a single-label pretrained model.
    \item \textbf{ReLabel}~\cite{relabel}: ReLabel uses patch-wise soft labels derived from the ReLabel maps to supervise random crops. We follow their training setup with cross-entropy loss.
    \item \textbf{ReLabel w/ Our Mask:} Our variant that applies ReLabel’s soft labels to our object proposals, padded with the original global label for fair comparison.
    \item \textbf{Ours:} Trained using our multi-label annotations as described in Sec.~\ref{sec:sweep_label}.
\end{itemize}

Training details are provided in Supp.~\ref{app:main_exp_config}. Results in Table~\ref{tab:main_exp} show that our method outperforms all baselines across most evaluation metrics. In particular, while single-positive learning methods like LL improve over the standard baseline on ReaL ($+0.2$ top-1 accuracy and +$0.1$ mAP), our relabeled supervision achieves consistently higher gains without introducing additional training complexity ($+1.6$ top-1 accuracy and +$1.1$ mAP on ReaL), and achieving the highest scores on ReaL ($88.2$), IN-Seg ($88.8$), and INv2-ML ($76.2$) in mAP.
While ReLabel achieves the highest top-1 accuracy on the original IN-Val ($78.9$), due to its soft-label optimization for the original single-label target, our method leads in all multi-label benchmarks, both in terms of top-1 accuracy and mAP. On average across the three multi-label datasets, our method improves mAP by $0.77$ and top-1 accuracy by $0.97$ over ReLabel.
Further, replacing ReLabel’s soft patch supervision with our localized multi-label targets improves performance across most multi-label benchmarks (e.g., $+0.4$, $+0.5$, and $+0.3$ accuracy on ReaL, IN-Seg, and INv2-ML, respectively), confirming the value of spatially grounded labels over soft distributions constrained to sum to one. A subgroup analysis in Table~\ref{tab:main_subgroup_real} further supports these findings: for images with multiple objects, our method yields an average of $+3.35$ mAP over single-label training baseline and $+1.48$ mAP over ReLabel. These results validate the benefit of explicit multi-label training, especially for complex real-world scenes.

\input{tables/in21k_mul_comparison}

\subsection{Robustness and Transferability}
\label{sec:arch_size_transfer}
We evaluate whether the benefits of our multi-label supervision generalize across architectures and support stronger transfer learning. All experiments in this section are conducted with an input size of $224$. Table~\ref{tab:cross_arch_size_224} presents results for five models: ResNet-50/101 and ViT-small/base/large. We explore two training modes: (1) end-to-end training from scratch using our multi-labels, and (2) fine-tuning a model pretrained on standard single-label ImageNet for $20$ epochs using our labels. The latter provides a practical and lightweight approach for improving off-the-shelf models.
For ResNet, we tune the hyperparameter to find the best setup of training original label with BCE Loss, and apply it directly to our multi-labels. For ViTs, we adopt the DeiT-3 training recipe~\cite{touvron2022deit}, which is already robust under BCE loss. Full training configurations are provided in Supp.~\ref{app:cross_arch_size_exp_config}.
We also evaluate downstream transfer by fine-tuning the pretrained models on multi-label classification benchmarks: Pascal VOC 2007 and MS COCO 2017.

\noindent \textbf{In-domain gains from fine-tuning.} 
Fine-tuning single-label pretrained models with our multi-label supervision consistently improves performance across evaluation sets. We observe top-1 accuracy gains of up to $+1.90$ on IN-ReaL and $+2.42$ on INv2-ML (both with ResNet-101), and $+1.24$ on INv2 (ResNet-50). Multi-label mAP improvements are also strong, with gains up to $+1.93$ on IN-ReaL and $+5.04$ on INv2-ML. These results show that fine-tuning with our labels is a lightweight and effective way to improve in-domain performance—without retraining from scratch.
Interestingly, improvements on multi-label benchmarks do not always correlate with gains on the original single-label IN-Val set. For example, ViT-base and ViT-large see no improvement or even a slight drop in IN-Val top-1 accuracy after fine-tuning, yet still show consistent gains across all multi-label benchmarks. This discrepancy highlights the limitations of single-label evaluation and underscores the value of richer, multi-label supervision.

\noindent \textbf{End-to-end training with multi-labels.} 
Training models from scratch using our multi-label annotations also consistently improves over single-label training. We observe top-1 accuracy gains of up to $+1.98$ on IN-ReaL (ResNet-101) and $+1.50$ on INv2 (ViT-small), along with mAP improvements up to $+1.91$ (ResNet-101, IN-ReaL) and $+5.15$ (ViT-small, INv2-ML).
Comparing end-to-end training with fine-tuning reveals useful trends: for smaller models like ViT-small, full training outperforms fine-tuning (e.g., $+0.4$ and $+1.4$ top-1 accuracy on IN-ReaL and INv2, respectively). For larger models such as ViT-base and ViT-large, fine-tuning offers comparable or slightly better gains. We hypothesize two contributing factors: (1) current hyperparameters—optimized for single-label—may not be ideal for multi-label training; and (2) larger models may require longer training to fully benefit from richer supervision.

\noindent \textbf{Transfer learning performance.} 
Our approach improves downstream multi-label transfer across all architectures: fine-tuning with our labels yields an average mAP gain of $+1.0$ on VOC and COCO, while end-to-end multi-label pretraining provides even larger gains ($+2.0$ on COCO, $+1.7$ on VOC). These results challenge the standard pipeline of single-label pretraining followed by multi-label fine-tuning~\cite{ridnik2021asymmetric,liu2021query2label,wang2021can}, showing that richer supervision from the outset yields stronger representations.
We hypothesize that multi-label training reduces representation collapse by encouraging more diverse features. Following~\cite{harun2025controlling}, we evaluate feature entropy and confirm that multi-label supervision consistently produces higher entropy than single-label training—supporting its benefit for out-of-distribution generalization (see Supp.~\ref{app:koleo}, Table~\ref{tab:koleo}).
Additionally, in Supp.~\ref{app:cross_arch_size_full} (Table~\ref{tab:cross_arch_size}), we show that ViTs trained with input size $384$ follow the same trends, confirming the generality of our multi-label method across models and image resolutions.

\subsection{Comparison with Semantic Multi-Label}
We compare our approach to MIIL~\cite{ridnik2021imagenet}, which constructs hierarchical semantic multi-labels from ImageNet-21K for pretraining, followed by fine-tuning on downstream tasks—including ImageNet-1K using standard single-label supervision (see Fig.~\ref{fig:example}). As shown in Table~\ref{tab:in21k_mul_comparison}, MIIL achieves strong results under this setup. However, when further fine-tuned with our explicit object-centric multi-label annotations, performance improves consistently across all multi-label ImageNet evaluation sets. 
Concretely, our approach improves top-1 accuracy by $+0.2$, $+0.4$, and $+0.5$, and mAP by $+0.1$, $+0.4$, and $+1.9$ on ReaL, IN-Seg, and INv2-ML, respectively.
Moreover, our method—trained end-to-end from scratch on ImageNet-1K using DeiT~\cite{touvron2022deit} training recipes and our multi-label supervision—matches or exceeds the performance of MIIL, despite MIIL relying on ImageNet-21K pretraining.
Notably, we observe stronger transfer performance on downstream benchmarks, with improvements of $+1.9$ mAP on COCO and $+2.4$ mAP on VOC. 
We attribute these gains to our label definitions, which emphasize concrete object presence and align more closely with real-world multi-label tasks than MIIL’s semantic hierarchies.

\subsection{Additional Experiments}
We present several additional analyses in Supp.~\ref{app:additional_exp} that demonstrate the broader utility of our learned region-level classifier. First, we show that training with soft spatial label maps derived from our classifier achieves comparable or better performance on ImageNet than ReLabel~\cite{relabel}, with both methods outperforming standard single-label training by $+1.4$ top-1 accuracy on IN-Val and ReaL. Second, we apply our classifier as a post-hoc filter to CutLER~\cite{wang2023cut} mask proposals, removing $12\%$ of noisy masks and improving the quality of pseudo-labels for unsupervised segmentation. Finally, we showcase an interactive labeling tool powered by our classifier, which enables fast and accurate annotation of arbitrary image regions. Together, these findings highlight the versatility and effectiveness of our object-centric classification head beyond standard classification tasks.

\section{Discussion and Conclusion}
This work revisits the foundational ImageNet-1K dataset and shows that its supervision can be substantially strengthened through a fully automated pipeline that produces explicit, region-grounded multi-label annotations. By identifying multiple object instances per image, our approach addresses the long-standing limitations of single-label supervision and delivers consistent gains across architectures, training regimes, and downstream multi-label tasks. Beyond accuracy, our annotations provide interpretable, proposal-level grounding that complements human verification and supports scalable dataset auditing.

More broadly, our results suggest that legacy datasets need not remain static: automated relabeling offers a practical path for continuously improving supervision quality at scale, with potential benefits for detection, multimodal grounding, and representation learning in future foundation models. The resulting labels also expose richer object co-occurrence patterns, enabling new research directions in bias analysis, compositional learning, and semi-automated annotation workflows.

While our method assumes one label per region—an occasional limitation for overlapping or hierarchical classes—we outline mitigation strategies using co-occurrence priors (Supp.~\ref{app:fix_ambiguous_class}). Further gains may be achieved with optimized configurations for larger models. We will release our code and annotations to foster research in multi-label learning, region-aware supervision, and automated dataset construction.

\paragraph{Acknowledgments.} This work was supported in part by NSF award \#2326491. The views and conclusions contained herein are those of the authors and should not be interpreted as representing the official policies or endorsements of any sponsor.

{
    \small
    \bibliographystyle{ieeenat_fullname}
    \bibliography{main}
}

\input{supplemental}

\end{document}

%% file: preamble.tex
\usepackage{multirow}
\usepackage[normalem]{ulem}
\useunder{\uline}{\ul}{}
\usepackage{adjustbox}
\usepackage{bm}
\usepackage{colortbl}
\usepackage{hhline}
\usepackage{booktabs}
\arrayrulecolor{black}
\usepackage{verbatim}

%% file: figures/relabel_example.tex
\begin{figure}[t]
    \centering
    \includegraphics[width=0.49\textwidth]{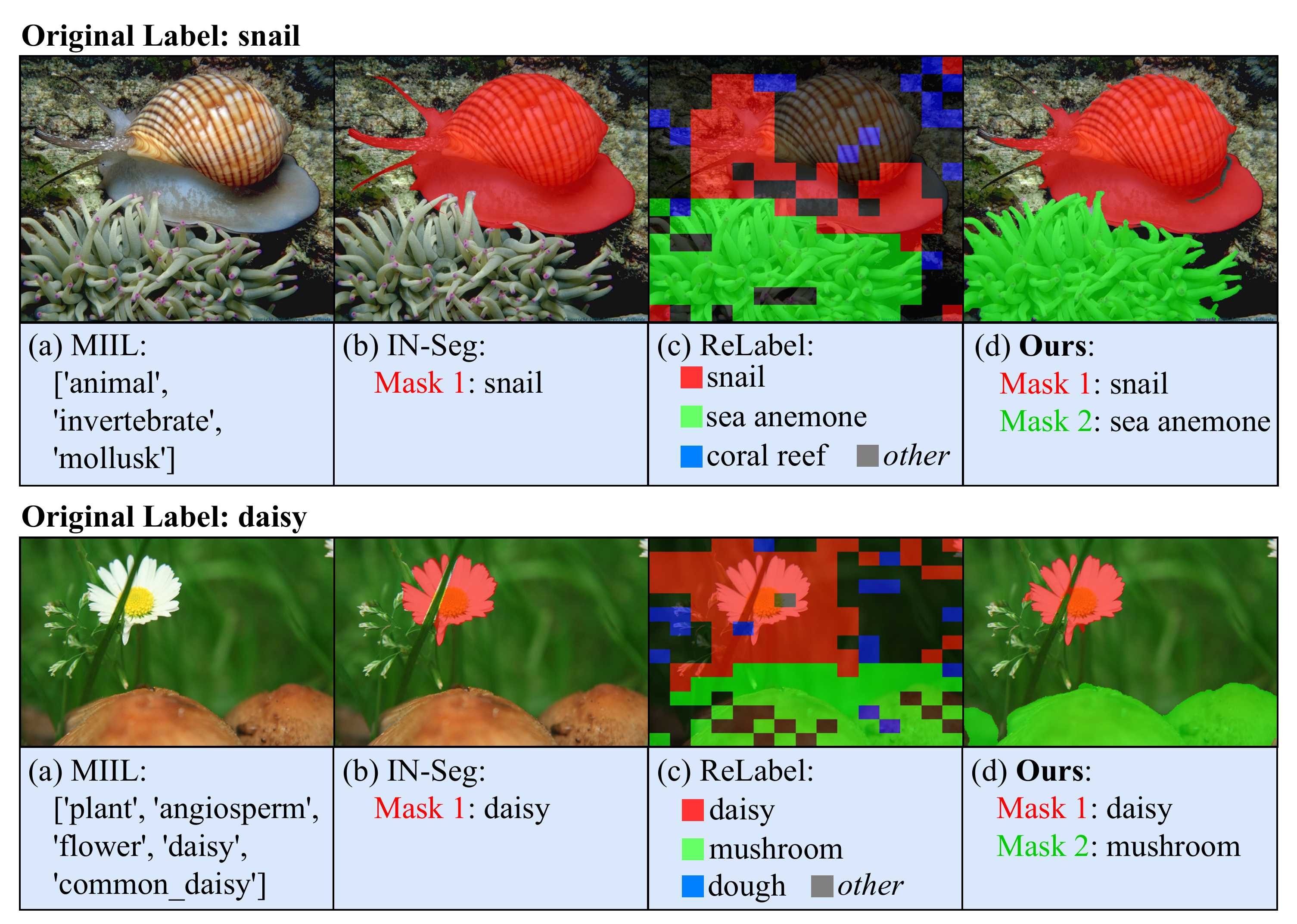}
    \vspace{-5mm}
    \caption{Comparison of Existing ImageNet Train-split Relabeling Strategies with Ours.
Original ImageNet~\cite{imagenet} annotations assume a single label per image. \textbf{(a)} MIIL~\cite{ridnik2021imagenet} adds hierarchical labels from ImageNet-21K but lacks object-level distinctions. \textbf{(b)} ImageNet-Segments~\cite{gao2022large} (IN-Seg) offers pixel masks for 9k training images with single object annotation. \textbf{(c)} ReLabel~\cite{relabel} assigns soft labels via a $15^2$ patch map, requiring crop coordinates to extract local supervision. \textbf{(d)} In contrast, our method generates explicit multi-label annotations with corresponding spatial masks, offering true multi-object labeling for the entire training set. 
}
\vspace{-5mm}
\label{fig:example}
\end{figure}

%% file: figures/diagram.tex
\begin{figure*}[t]
    \centering
    \includegraphics[width=1.0\textwidth]{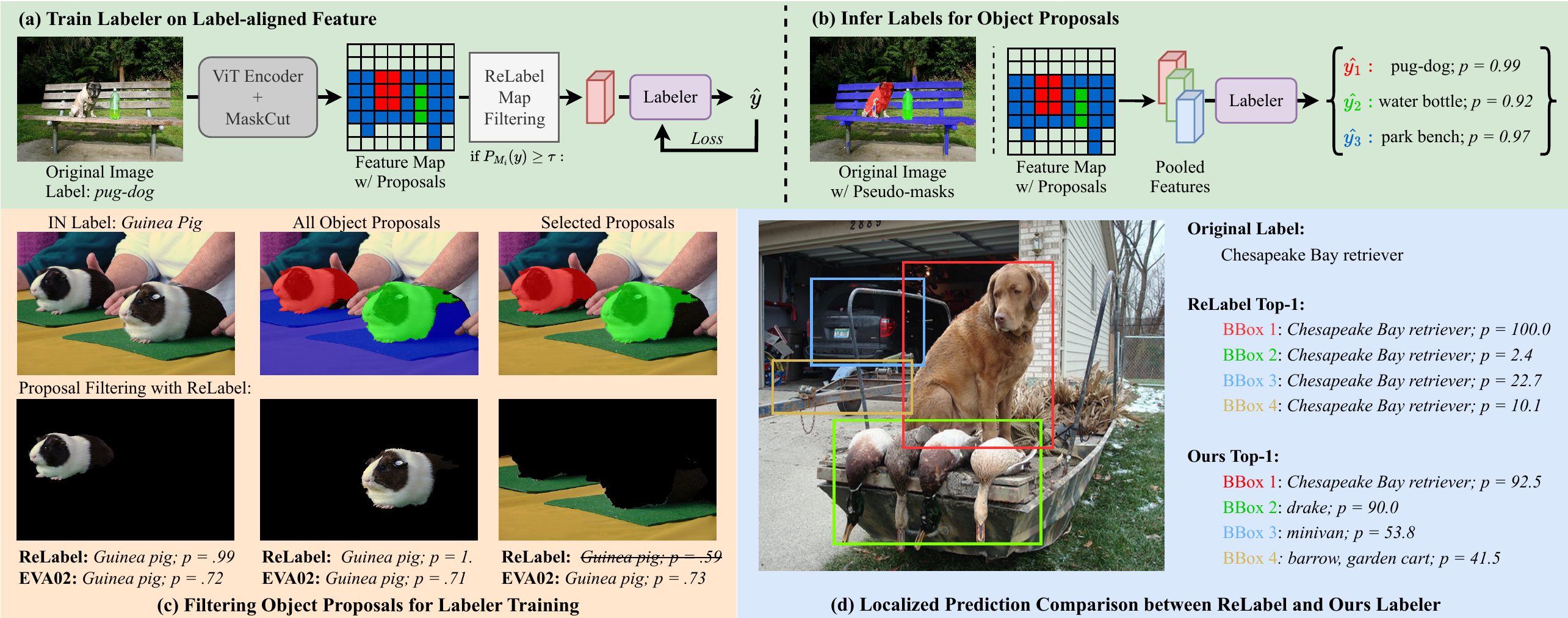}
\vspace{-7mm}
    \caption{Overview of our relabeling pipeline. \textbf{(a)} We apply MaskCut~\cite{wang2023cut} on DINOv3~\cite{simeoni2025dinov3} ViT features to generate object proposals. ReLabel~\cite{relabel} maps are used to filter proposals most aligned with the original ground-truth label, which supervise a lightweight labeler. \textbf{(b)} At inference, the labeler predicts class scores for each proposal, enabling spatially grounded multi-label annotations. 
    \textbf{(c)} Compared to a global classifier (e.g., EVA02~\cite{fang2024eva}), ReLabel improves proposal filtering while can still produce high-confidence false positives. \textbf{(d)} Visualization of top-1 predictions per region shows our labeler better disambiguates multiple objects than ReLabel, avoiding context bias and recognizing distinct object categories. 
    }
\vspace{-5mm}
    \label{fig:diagram}
\end{figure*}

%% file: figures/qualitative.tex
\begin{figure*}[t]
    \centering
    \includegraphics[width=0.99\textwidth]{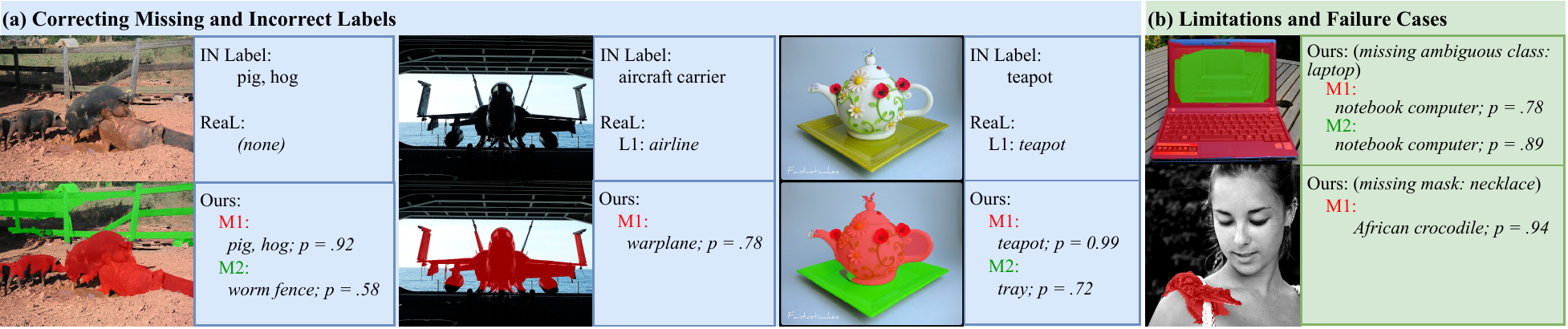}
    \vspace{-3mm}
    \caption{Qualitative examples comparing our multi-label annotations against ImageNet and ReaL~\cite{real}.
\textbf{(a)} Our method successfully corrects missing or incorrect labels from ReaL by identifying additional objects and providing improved grounding.
\textbf{(b)} Representative failure cases, including ambiguity (e.g., notebook vs. laptop) and missed object proposals.}
\vspace{-5mm}
    \label{fig:qualitative}
\end{figure*}

%% file: tables/main_exp.tex
\begin{table*}[t]
\caption{Top‑1 accuracy (\%) on original ImageNet (IN) and ImageNet‑V2 (INv2), and mAP (\%) on multi‑label validation sets (ReaL, IN-Seg
and INv2‑ML) for ResNet‑50. The best and 2nd best performances are highlighted by bold and underline, respectively. Our method consistently outperforms single‑label training and single‑positive methods. 
}
\vspace{-3mm}
\label{tab:main_exp}
\centering
\scriptsize
\resizebox{0.8\textwidth}{!}{
\begin{tabular}{l|ccccc|ccc}
\hline
\multicolumn{1}{c|}{\multirow{2}{*}{\textbf{Method}}} &
  \multicolumn{5}{c|}{\textbf{Top-1 Acc $\uparrow$}} &
  \multicolumn{3}{c}{\textbf{Multi-Label: mAP $\uparrow$}} \\ \cline{2-9} 
\multicolumn{1}{c|}{} &
\textbf{IN-Val} & 
\textbf{ReaL} & 
\textbf{IN-Seg} & 
\textbf{IN-v2} & 
\textbf{INv2-ML} &
\textbf{ReaL} & 
\textbf{IN-Seg} & 
\textbf{INv2-ML} \\
 \hline
Original Label          & 77.6          & 84.0       & 84.3       & 65.4       & 77.4       & 87.1       & 87.8       & 73.0       \\
Original + Label Smooth & 78.2          & 84.1       & 84.4       & 66.1       & 78.2       & 87.0       & 87.7       & 72.3       \\
SCL~\cite{verelst2023spatial} (reported)  & 76.9          & 83.4       & -          & -          & -          & 82.2       & -          & -          \\
LL~\cite{kim2022large}             & 77.8          & 84.2       & 84.3       & 65.7       & 77.7       & 87.2       & 87.7       & 72.7       \\
ReLabel~\cite{relabel}                & \textbf{78.9} & 85.0       & 84.8       & {\ul 67.3} & 79.4       & 87.9       & 88.2       & {\ul 74.8} \\
ReLabel~\cite{relabel} w/ Our Mask    & {\ul 78.8}    & {\ul 85.4} & {\ul 85.3} & 67.2       & {\ul 79.7} & {\ul 88.0} & {\ul 88.5} & 74.6       \\
\textbf{Multi-label (Ours)} &
  78.7  &
  \textbf{85.6} &
  \textbf{85.5} &
  \textbf{67.4} &
  \textbf{81.0} &
  \textbf{88.2} &
  \textbf{88.8} &
  \textbf{76.2} \\ \hline
\end{tabular}
}
\vspace{-5mm}
\end{table*}

%% file: tables/main_exp_subgroup_real.tex
\begin{table}[t]
\caption{Subgroup analysis of multi-label classification performance on ReaL. We report mAP overall and stratified by number of ground-truth labels (k) per image. The best and 2nd best performances are highlighted by bold and underline, respectively.  
}
\vspace{-3mm}
\label{tab:main_subgroup_real}
\centering
\scriptsize
\resizebox{0.48\textwidth}{!}{
\begin{tabular}{l|ccccc}
\hline
\multicolumn{1}{c|}{\multirow{3}{*}{\textbf{Method}}} & \multicolumn{5}{c}{\textbf{ReaL: mAP $\uparrow$}}                                          \\ \cline{2-6} 
\multicolumn{1}{c|}{} & \textbf{k=all:} & \textbf{k=1:} & \textbf{k=2:} & \textbf{k=3:} & \textbf{k}\bm{$\ge$}\textbf{4:} \\
\multicolumn{1}{c|}{}                                         & \textbf{46837} & \textbf{39394} & \textbf{5408} & \textbf{1319} & \textbf{716}  \\ \hline
Original Label                                                & 87.1           & 90.6           & 71.1          & 61.7          & 60.8          \\
Original + Label Smooth                                       & 87.0           & 90.5           & 70.8          & 60.3          & 59.7          \\
SCL~\cite{verelst2023spatial} (reported)                                                & 82.2           & 88.8           & 61.4          & 36.8          & 21.1          \\
LL~\cite{kim2022large}                                                            & 87.2           & 90.7           & 71.4          & 61.4          & 61.5          \\
ReLable~\cite{relabel}                                                       & 87.9           & 91.2           & 73.2          & 63.1          & {\ul 62.9}    \\
ReLable~\cite{relabel} w/ Our mask                                           & {\ul 88.0}     & {\ul 91.2}     & {\ul 73.5}    & {\ul 63.7}    & 62.5          \\
Multi-label (Ours)                                            & \textbf{88.2}  & \textbf{91.3}  & \textbf{74.3} & \textbf{65.3} & \textbf{64.0} \\ \hline
\end{tabular}
}
\vspace{-4mm}
\end{table}

%% file: tables/cross_arch_size_224.tex
\begin{table*}
\setlength{\arrayrulewidth}{0.6pt}
\arrayrulecolor{black} 
\caption{End-to-end training and transfer performance with our multi-label annotations. We compare single-label training (Single-label E2E), fine-tuning with our multi-labels (+ Multi-label FT), and end-to-end multi-label training (Multi-label E2E) across various model architectures. 
Best results are highlighted in bold. Our multi-label supervision improves in-domain performance on ImageNet and its variants, and yields consistent gains in downstream multi-label transfer to COCO and VOC. 
}
\vspace{-3mm}
\label{tab:cross_arch_size_224}
\centering
\scriptsize
\begin{tabular}{cccccccccccc}
\toprule
\multicolumn{2}{c}{} & \multicolumn{8}{c}{\textbf{ImageNet Evaluation}} & \multicolumn{2}{c}{\textbf{Transfer Learning}} \\
\cmidrule(lr){3-10} \cmidrule(lr){11-12}
\textbf{Model} & \textbf{Method} & \multicolumn{5}{c}{\textbf{Top-1 Acc $\uparrow$}} & \multicolumn{3}{c}{\textbf{Multi-Label: mAP $\uparrow$}} & \multicolumn{2}{c}{\textbf{mAP $\uparrow$}} \\
\cmidrule(lr){3-7} \cmidrule(lr){8-10} \cmidrule(lr){11-12}
 & & \textbf{IN-Val} & \textbf{ReaL} & \textbf{IN-Seg} & \textbf{INv2} & \textbf{INv2-ML} & \textbf{ReaL} & \textbf{IN-Seg} & \textbf{INv2-ML} & \textbf{COCO} & \textbf{VOC} \\
\hline
\rowcolor[HTML]{F2F2F2} 
\cellcolor[HTML]{F2F2F2} &
  Single-label E2E &
  78.2 &
  84.1 &
  84.4 &
  66.1 &
  78.2 &
  87.0 &
  87.7 &
  72.3 &
  77.0 &
  89.2 \\
\rowcolor[HTML]{E3F2D9} 
\cellcolor[HTML]{F2F2F2} &
  +Multi-label FT &
  \textbf{78.7} &
  85.4 &
  \textbf{85.9} &
  67.3 &
  80.4 &
  88.2 &
  \textbf{89.1} &
  76.1 &
  78.7 &
  90.5 \\
\rowcolor[HTML]{C9E4B4} 
\multirow{-3}{*}{\cellcolor[HTML]{F2F2F2}ResNet-50} &
  Multi-label E2E &
  78.7 &
  \textbf{85.6} &
  85.5 &
  \textbf{67.4} &
  \textbf{81.0} &
  \textbf{88.2} &
  88.8 &
  \textbf{76.2} &
  \textbf{78.9} &
  \textbf{90.7} \\\hline
\rowcolor[HTML]{F2F2F2} 
\cellcolor[HTML]{F2F2F2} &
  Single-label E2E &
  79.3 &
  84.7 &
  85.3 &
  68.2 &
  80.0 &
  87.1 &
  88.2 &
  72.7 &
  79.5 &
  90.4 \\
\rowcolor[HTML]{E3F2D9} 
\cellcolor[HTML]{F2F2F2} &
  +Multi-label FT &
  \textbf{80.3} &
  86.6 &
  \textbf{87.1} &
  69.0 &
  \textbf{82.4} &
  89.0 &
  \textbf{90.1} &
  \textbf{77.7} &
  80.1 &
  91.2 \\
\rowcolor[HTML]{C9E4B4} 
\multirow{-3}{*}{\cellcolor[HTML]{F2F2F2}ResNet-101} &
  Multi-label E2E &
  80.2 &
  \textbf{86.7} &
  86.9 &
  \textbf{69.2} &
  82.3 &
  \textbf{89.0} &
  89.9 &
  77.4 &
  \textbf{80.2} &
  \textbf{92.2} \\\hline
\rowcolor[HTML]{F2F2F2} 
\cellcolor[HTML]{F2F2F2} &
  Single-label E2E &
  81.4 &
  87.0 &
  87.2 &
  70.7 &
  83.1 &
  89.0 &
  89.7 &
  75.6 &
  79.1 &
  91.0 \\
\rowcolor[HTML]{E3F2D9} 
\cellcolor[HTML]{F2F2F2} &
  +Multi-label FT &
  81.5 &
  87.7 &
  87.8 &
  70.8 &
  84.1 &
  90.1 &
  90.9 &
  80.2 &
  81.0 &
  91.9 \\
\rowcolor[HTML]{C9E4B4} 
\multirow{-3}{*}{\cellcolor[HTML]{F2F2F2}ViT-small} &
  Multi-label E2E &
  \textbf{82.0} &
  \textbf{88.1} &
  \textbf{88.3} &
  \textbf{72.2} &
  \textbf{85.1} &
  \textbf{90.3} &
  \textbf{91.1} &
  \textbf{80.7} &
  \textbf{83.3} &
  \textbf{93.3} \\\hline
\rowcolor[HTML]{F2F2F2} 
\cellcolor[HTML]{F2F2F2} &
  Single-label E2E &
  \textbf{83.7} &
  88.2 &
  88.2 &
  73.6 &
  85.8 &
  90.6 &
  90.9 &
  80.3 &
  83.0 &
  92.7 \\
\rowcolor[HTML]{E3F2D9} 
\cellcolor[HTML]{F2F2F2} &
  +Multi-label FT &
  83.6 &
  \textbf{88.9} &
  88.8 &
  \textbf{74.0} &
  \textbf{86.9} &
  90.6 &
  91.3 &
  81.3 &
  83.8 &
  93.6 \\
\rowcolor[HTML]{C9E4B4} 
\multirow{-3}{*}{\cellcolor[HTML]{F2F2F2}ViT-base} &
  Multi-label E2E &
  83.4 &
  88.8 &
  \textbf{89.1} &
  73.9 &
  86.8 &
  \textbf{90.7} &
  \textbf{91.5} &
  \textbf{81.8} &
  \textbf{84.7} &
  \textbf{94.5} \\\hline
\rowcolor[HTML]{F2F2F2} 
\cellcolor[HTML]{F2F2F2} &
  Single-label E2E &
  84.6 &
  88.6 &
  88.5 &
  74.7 &
  87.1 &
  90.8 &
  91.2 &
  81.4 &
  84.8 &
  93.4 \\
\rowcolor[HTML]{E3F2D9} 
\cellcolor[HTML]{F2F2F2} &
  +Multi-label FT &
  \textbf{84.6} &
  \textbf{89.3} &
  \textbf{89.4} &
  \textbf{75.1} &
  \textbf{88.2} &
  91.1 &
  91.9 &
  \textbf{83.3} &
  85.2 &
  94.4 \\
\rowcolor[HTML]{C9E4B4} 
\multirow{-3}{*}{\cellcolor[HTML]{F2F2F2}ViT-large} &
  Multi-label E2E &
  84.3 &
  89.3 &
  89.4 &
  74.9 &
  87.9 &
  \textbf{91.2} &
  \textbf{91.9} &
  83.0 &
  \textbf{86.4} &
  \textbf{95.0} \\\hline

\end{tabular}
\scriptsize
\vspace{-4mm}
\end{table*}

%% file: tables/in21k_mul_comparison.tex
\begin{table*}[t]
\caption{Comparison with MIIL~\cite{ridnik2021imagenet}, which uses hierarchical multi-labels from ImageNet-21K for pretraining. While MIIL fine-tunes on ImageNet-1K (IN1k) using single-label (Sig) supervision, fine-tuning MIIL with our multi-labels (Mul) improves all multi-label benchmarks. our end-to-end multi-label (Mul E2E) training—without 21K pretraining—achieves comparable in-domain performance and better downstream transfer results. Results use \texttt{ViT-B/16} at $224$ resolution. Best results are highlighted in bold.
}
\vspace{-3mm}
\label{tab:in21k_mul_comparison}
\centering
\resizebox{0.85\textwidth}{!}{
\begin{tabular}{ccccccccccc}
\toprule
\multicolumn{1}{c}{} & \multicolumn{8}{c}{\textbf{ImageNet Evaluation}} & \multicolumn{2}{c}{\textbf{Transfer Learning}} \\
\cmidrule(lr){2-9} \cmidrule(lr){10-11}
\textbf{Method} & \multicolumn{5}{c}{\textbf{Top-1 Acc $\uparrow$}} & \multicolumn{3}{c}{\textbf{Multi-Label: mAP $\uparrow$}} & \multicolumn{2}{c}{\textbf{mAP $\uparrow$}} \\
\cmidrule(lr){2-6} \cmidrule(lr){7-9} \cmidrule(lr){10-11}
 & \textbf{IN-Val} & \textbf{ReaL} & \textbf{IN-Seg} & \textbf{INv2} & \textbf{INv2-ML} & \textbf{ReaL} & \textbf{IN-Seg} & \textbf{INv2-ML} & \textbf{COCO} & \textbf{VOC} \\
\midrule
IN21k-Mul Pretrain    & -    & -    & -    & -    & \multicolumn{1}{c}{-}    & -    & -    & -    & 82.8 & 92.1 \\
+IN1k-Sig FT    & \textbf{84.3} & 88.7 & 88.8 & \textbf{74.0} & \multicolumn{1}{c}{86.4} & 90.9 & 91.4 & 80.7 & 82.3 & 90.5 \\
+IN1k-Mul FT    & 83.6 & \textbf{88.9} & \textbf{89.2} & 73.8 & \multicolumn{1}{c}{\textbf{86.9}} & \textbf{91.0} & \textbf{91.8} & \textbf{82.6} & 82.5 & 90.9 \\ \hline
IN1k-Mul E2E & 83.4 & 88.8 & 89.1 & 73.9 & \multicolumn{1}{c}{86.8} & 90.7 & 91.5 & 81.8 & \textbf{84.7} & \textbf{94.5} \\ \hline
\end{tabular}
}
\vspace{-4mm}
\end{table*}

%% file: supplemental.tex
\clearpage
\setcounter{page}{1}
\maketitlesupplementary

\appendix

\noindent We organize our supplementary material as follows:
\begin{itemize}
    \item Supp.~\ref{app:mcut} details our object proposal generation pipeline, including MaskCut implementation, hyperparameter tuning, and comparison to SAM.
    \item Supp.~\ref{app:labeler_detail} outlines the labeler training setup, relabeling statistics, and analysis of different label aggregation strategies.
    \item Supp.~\ref{app:main_exp} provides experimental protocols for ImageNet training and transfer learning across architectures and input sizes.
    \item Supp.~\ref{app:human_comparison} presents a qualitative comparison between our labels and human-curated ReaL annotations.
    \item Supp.~\ref{app:fix_ambiguous_class} details ambiguous class definitions in ImageNet and proposes two strategies to mitigate label noise.
    \item Supp.~\ref{app:additional_exp} contains additional analyses, including resolution robustness, feature entropy, soft-label training, mask filtering for CutLER, and our interactive annotation tool.
\end{itemize}

\section{Object Proposal Generation}
\label{app:mcut}
\subsection{MaskCut Implementation Details}
\label{app:mcut_detail}
This section details MaskCut~\cite{wang2023cut}, which we use to extract object proposals from an image using a self-supervised vision transformer (ViT). The goal is to discover multiple salient object regions per image $I \in \mathbb{R}^{3 \times h \times w}$ without any manual labels. 
MaskCut builds on the idea of iterative normalized cuts, generating multiple binary masks that segment distinct object-like regions. Let $\mathcal{F}$ be a pretrained ViT encoder. Given an input image $I$, we first extract its patch-level feature map $F = \mathcal{F}(I) \in \mathbb{R}^{h' \times w' \times d}$, which is flattened into $N = h' \times w'$ feature vectors ${f_1, f_2, \dots, f_N}$ of dimension $d$.
We then construct a fully connected graph over the $N$ patches, where the affinity between two nodes $i$ and $j$ is defined by the cosine similarity of their features:
\begin{equation}
    W_{ij} = \frac{f_i \cdot f_j}{\lVert f_i \rVert_2 \, \lVert f_j \rVert_2}
\end{equation}
Normalized Cuts (NCut) is applied to partition the graph into foreground vs. background. NCut finds an eigenvector $x \in \mathbb{R}^N$ (the relaxed indicator of the cut) by solving the generalized eigenvalue problem
\begin{equation}
    (D-W)x = \lambda Dx
\end{equation}
and we take the eigenvector $x$ corresponding to the second-smallest eigenvalue $\lambda$ (standard practice for NCut). We then threshold $x$ to produce an initial binary mask $M$ for the foreground object:
\begin{equation}
    M(i) =
    \begin{cases}
    1, & x(i) \ge \mu(x), \\
    0, & x(i) < \mu(x),
    \end{cases}
\end{equation}
where $\mu(x)$ is the mean value of $x$. Here $M(i)=1$ indicates patch $i$ is classified as foreground.
We determine which side of the cut is the object (foreground) using two criteria from MaskCut~\cite{wang2023cut}: (a) the foreground mask should contain the patch corresponding to the largest magnitude in $x$ (since the principal object tends to dominate the eigenvector), and (b) the foreground should not include more than one or two image corner patches (to avoid selecting the entire background as object). If our initial mask $M$ does not satisfy these criteria (e.g. the largest-eigenvector patch is not in $M$, or $M$ covers too many corners), we flip the assignment (set $M \leftarrow 1 - M$). This ensures $M$ corresponds to a salient object in the image. 
Following MaskCut, we also  threshold the affinity matrix by a hyperparameter $\tau^{\text{ncut}}$ to sharpen the segmentation by setting all $W_{ij} < \tau^{\text{ncut}}$ to $1e^{-5}$ and  $W_{ij} \ge \tau^{\text{ncut}}$ to $1$. A Conditional Random Field (CRF~\cite{sutton2012introduction}) post-processing step is then applied to the patch mask to incorporate low-level pixel continuity, yielding a refined pseudo segmentation mask for the object. To discover multiple objects, we iteratively repeat the NCut process on the remaining image regions. After obtaining the first object mask $P_1$ (the set of patches with $M(i)=1$), we mask out those patches by removing them from the graph. In practice, we set their feature vectors to zero or exclude them from further similarity computations. Formally, let $U_1 = P_1$ be the set of foreground patches found. For the next iteration, we update the patch similarities so that any patch $i \in U_1$ (or patch $j \in U_1$) has no affinity:
\begin{equation}
    W^{(2)}_{ij} =
        \begin{cases}
        \dfrac{f_i \cdot f_j}{\lVert f_i \rVert_2 \, \lVert f_j \rVert_2}, & i \notin U_1 \text{ and } j \notin U_1, \\
        0, & \text{otherwise}.
        \end{cases}
\end{equation}
We then rerun NCut on this updated matrix $W^{(2)}$ to find the next object mask $P_2$. We continue this masked NCut procedure for $t=1,2,\dots,N$, each time excluding all patches from previously found masks $U_t = \bigcup_{s=1}^{t} P_s$. This yields up to $N$ object masks ${P_1, P_2, \dots, P_N}$ per image. We set the maximum iteration $N$ as a hyperparameter, which is the maximum number of object proposals produced for one image. The process stops early if an iteration returns no significant foreground (e.g. only background remains).

\subsection{Hyperparameter Selection}
\label{app:mcut_sweep}
The performance of MaskCut depends on several key hyperparameters: the affinity threshold $\tau^{\text{ncut}}$, the number of object proposals $N$, the choice of ViT backbone, and the use of CRF post-processing. A high $\tau^{\text{ncut}}$ may over-merge nearby objects (e.g., grouping multiple dogs), while a low value risks over-segmentation. Similarly, a small $N$ may miss valid objects, whereas a large $N$ can introduce spurious proposals. We explore various self-supervised ViT backbones: DINOv1~\cite{dino} (as used in the original MaskCut), DINOv2~\cite{oquab2023dinov2} (ViT-S/B/L/G), and DINOv3~\cite{simeoni2025dinov3} (ViT-S+/B/L/H). For each, we sweep across:
\begin{itemize}
\item Input image resolutions (corresponding to patch grids of $24 \times 24$, $32 \times 32$, or $48 \times 48$),
\item Feature type (last-layer patch features or last attention query/key/value),
\item $\tau^{\text{ncut}} \in [0.1, 0.9]$ in steps of $0.1$, followed by fine sweeps of $\pm0.05$ in $0.01$ increments,
\item Number of proposals $N \in \{3, 4, 5\}$,
\item CRF post-processing (enabled or disabled).
\end{itemize}

To evaluate, we compute object recall at IoU $\geq 0.5$ on $200$ randomly sampled images from ImageNet-Segments~\cite{gao2022large}. For DINOv1, we adopt the best configuration from MaskCut. For DINOv2 and DINOv3, we rank the top $7$ configurations by recall. We  then further manually assess visual quality on a $3$-point scale: (1) noisy/missing masks, (2) good with some issues, (3) all good masks. The top $4$ configurations with complementary strengths are selected for the final pipeline:
\begin{itemize}
\item \texttt{DINOv3-B/patch\_16/feat\_v}, input $768$, $\tau=0.35$, $N=4$, CRF off
\item \texttt{DINOv2-G/patch\_14/feat\_v}, input $672$, $\tau=0.12$, $N=3$, CRF on
\item \texttt{DINOv2-L/patch\_14/feat\_v}, input $448$, $\tau=0.12$, $N=3$, CRF on
\item \texttt{DINOv1-B/patch\_8/feat\_k}, input $480$, $\tau=0.15$, $N=3$, CRF on
\end{itemize}
We find that no single configuration is optimal for all images; using a diverse ensemble of top-performing setups improves robustness and overall coverage.

\subsection{Comparison with SAM}
\label{app:sam}
\input{figures/sam_example}
We evaluated Segment Anything (SAMv2~\cite{ravi2024sam}) as a potential object proposal generator by testing different automatic prompt configurations. As shown in Fig.~\ref{fig:sam}, SAM can produce detailed masks in some cases (e.g., config.2) but struggles with consistency—often under-segmenting, over-segmenting, or missing objects entirely in others. This variability across images makes it difficult to identify a fixed set of configurations that work reliably across the diverse ImageNet distribution. In contrast, MaskCut offers more stable and interpretable object-level proposals with consistent hyperparameters, making it better suited for our large-scale relabeling pipeline.

\section{Labeler Training Details}
\label{app:labeler_detail}
\subsection{Training Setup and Hyperparameters}
\label{app:labeler_training_details}
We filter object proposals by requiring the ReLabel confidence on the image’s original label to exceed a threshold of $\tau_{\text{sel}} = 0.75$.
The classification head is trained on input images resized to $512 \times 512$ resolution for $300$ epochs, using SGD with Nesterov momentum ($0.9$), weight decay of $1e^{-4}$, and learning rate of $0.1$. We adopt a cosine learning rate schedule with $5$ epochs of linear warm-up. The backbone (\texttt{DINOv3 ViT-L/16}) remains frozen throughout training. The global batch size is $512$.
We apply standard data augmentations including RandomResizedCrop, horizontal flip, and RandAugment~\cite{cubuk2020randaugment}. All geometric augmentations are applied consistently to both the original image and its associated object proposal to preserve spatial alignment.
To improve robustness, we introduce patch-level dropout: for each proposal mask $P$, we randomly drop $25\%$ of active (i.e., foreground) patches prior to feature pooling.
For reference, the trained labeler achieves a top-1 accuracy of $84.73\%$ on IN-Val and $88.61\%$ on ReaL. Note that this differs from the relabeling stage, where predictions are made by pooling over localized object regions rather than the entire patch map. These values are provided for completeness.

\subsection{Train-Set Relabeling Statistics}
\label{app:train_set_relabel_statistics}
\input{tables/train_statistics}
To better illustrate the label density in our relabeled ImageNet-1K training set, we apply a confidence threshold of $\tau = 0.5$ to the softmax outputs during inference to filter out low-confidence predictions. Note that this filtering is not used during training with soft labels. Table~\ref{tab:train_statistics} summarizes the distribution of predicted labels per image. Notably, $20.5\%$ of images have two or more labels, underscoring the multi-object nature of the dataset and the limitations of its original single-label annotations.

\input{tables/sweep_label_setup}
\subsection{Label Aggregation: Hard vs. Soft, Local vs. Global}
\label{app:sweep_label_setup}
We conduct a comprehensive evaluation of label aggregation strategies by training a ResNet-50 for $100$ epochs on ImageNet-1K using binary cross-entropy (BCE) loss. For a fair comparison with the original single label (i.e., converted to one-hot label) setup, we first sweep key hyperparameters and stabilization techniques known to affect BCE training~\cite{cui2019class,ridnik2021asymmetric,wightman2021resnet}. These include:
\begin{itemize}
    \item Positive label weighting (pos\_weight)
    \item Label smoothing which uses a relaxed (min, max) range instead of strict binary targets
    \item Head initialization to avoid early overconfidence (sigmoid outputs initialized near the smoothing minimum)
    \item Excluding the classification head from weight decay
    \item Optimizer: SGD vs. AdamW, with variations in learning rate and weight decay (WD)
\end{itemize}
The best configuration we identify is:
\textit{AdamW optimizer} with a learning rate of $0.001$ and weight decay of $0.15$ (WD=0 for the classifier head), a \textit{label smoothing range of $(0.0001, 0.95)$}, and a \textit{global batch size of $1024$}. Training uses cosine decay with $5$-epoch linear warmup, and standard augmentations  of RandomResizedCrop, horizontal flip, and RandAugment. Under this setup, BCE training with the original single-label supervision slightly outperforms the best cross-entropy configuration we found (IN-Val: $77.6$ vs. $77.5$). We then apply this setup across the label aggregation strategies described in Sec.~\ref{sec:sweep_label}. For \textit{Local-Hard}, we sweep $\tau$ from $0.3$ to $0.9$ and report the best-performing settings. Table~\ref{tab:sweep_label_setup} summarizes the performance across ImageNet benchmarks.

Several key observations emerge from the results. First, across nearly all configurations, our multi-label annotations outperform the original single-label supervision. The only exception is a marginal drop in IN-Val top-1 accuracy when no global label signal is used. Otherwise, we observe consistent improvements—up to $+0.6$ and $+1.1$ in top-1 accuracy on ReaL and INv2, and up to $+0.6$ and $+2.0$ mAP on ReaL and INv2-ML—demonstrating the effectiveness of our richer supervision. Among the aggregation strategies, combining local labels with a global signal consistently improves performance, especially on more challenging benchmarks like INv2 and INv2-ML. Soft labels also outperform hard labels across most metrics, supporting the value of preserving class confidence scores rather than applying fixed thresholds. Using the original ImageNet label as the global signal slightly outperforms our classifier’s own global prediction (e.g., $+0.2$ top-1 on IN-Val and ReaL). We attribute this to the classifier being optimized for localized regions, making it less superior for global image-level predictions. Overall, the best-performing configuration, \textit{Local-Soft + Original}, achieves the highest mAP on the challenging INv2-ML ($74.7$) and strong top-1 accuracy across all evaluation sets. Based on these findings, we adopt this setup as the default for all experiments in the main paper.

\section{Main Experiment Protocols}
\label{app:main_exp}
\subsection{Training Setup for ImageNet}
\label{app:main_exp_config}
For methods including \textit{Original Label} (with or without label smoothing), \textit{ReLabel~\cite{relabel}} with our object masks, and our \textit{Multi-label} supervision—we adopt the best BCE training setup identified in Supp.~\ref{app:sweep_label_setup}, extending training to $300$ epochs. For \textit{ReLabel}, we use the official released checkpoint (trained for $300$ epochs with CE loss).
For \textit{Large Loss} (LL~\cite{kim2022large}), we follow their procedure: fine-tuning from the label-smoothed baseline checkpoint for $10$ epochs, while sweeping key parameters: downweight ratio $\delta_\text{rel} \in [0.01, 0.4]$, revision mode {\texttt{LL-R}(\text{reject}), \texttt{LL-Ct}(\text{temporary correct}), \texttt{LL-Cp}(\text{permanent correct})}, and learning rate $\texttt{lr} \in [1e\text{-}5, 1e\text{-}4]$. The best-performing configuration is reported.
For \textit{SCL}~\cite{verelst2023spatial}, we report numbers from their original paper because code for SCL has not been released. Their method also finetunes an ImageNet-pretrained model using standard single-label supervision.

\subsection{Cross-Architecture Robustness and Transfer }
\label{app:cross_arch_size_exp_config}
\noindent \textbf{ResNet Training.} For all ResNet experiments (original and our multi-label variants), we use the same $300$-epoch BCE setup from Supp.~\ref{app:sweep_label_setup}.

\noindent \textbf{ViT Training.} For ViTs, we adopt the DeiT-III~\cite{touvron2022deit} training recipe, which is well-suited for BCE loss. We adapt their implementation by replacing mixup and cutmix from \texttt{timm} (which assumes single-label) with \texttt{torchvision} versions that support true multi-label training. All other hyperparameters remain unchanged. We compare three modes: (1) single-label training (using released checkpoints), (2) fine-tuning with our multi-labels (from released checkpoints), and (3) end-to-end multi-label training from scratch. We report the best performance under each metric with or without resize-center crop (scale ratio of $256 / 224$) during evaluation.

\noindent \textbf{Transfer Learning.} For transfer to VOC and COCO, we fine-tune each model for $100$ epochs. We attach a randomly initialized linear head of shape $(1000, N_c)$, where $N_c$ is the number of target classes ($20$ for VOC, $80$ for COCO). This head is folded into the classifier after training, introducing no overhead at inference and empirically enabling stable and fast convergence.
We use BCE loss with label smoothing $[0.001, 0.99]$, weight decay of $1e\text{-}4$, and sweep learning rates in $[5e\text{-}6,3e\text{-}4]$. The input resolution matches the pretraining resolution of each checkpoint. No center crop is used during evaluation. Best results are reported (in Table~\ref{tab:cross_arch_size_224},~\ref{tab:cross_arch_size}).

\input{figures/qualitative_appendix}
\section{Comparison with Human Annotations}
\label{app:human_comparison}
\subsection{Qualitative Breakdown of Human Agreement}
\label{app:qualitative}
We applied our relabeling pipeline to the ImageNet validation set and compared the results against ReaL~\cite{real}, the human-verified multilabel annotations. ReaL was constructed through a rigorous procedure: a diverse pool of $19$ trained models was used to propose candidate labels for each image (including each model’s top-1 prediction, other high-confidence predictions, and the image's original label), which was then narrowed to 6 models to ensure $>97\%$ recall of true labels (i.e., utilizing a subset of the validation set labeled by experts as the golden standard). Human annotators reviewed on average $7$–$8$ proposed labels per image, voting whether each label was present. Notably, if all $6$ models agreed on the original ImageNet label, the image was not re-annotated by humans (this happened for $25,111$ images). The final ReaL set contains $57,553$ labels spanning $46,837$ images, leaving $3,163$ images with no label after this process. Here we assess how our pipeline’s outputs compare to it.

To structure the comparison, we convert the label set into one-hot fashion by thresholding softmax score of $0.5$ \footnote{While tuning the threshold could increase the exact-match ratio with ReaL labels to as high as $70\%$, we use a fixed threshold of $0.5$ to balance precision and recall.}, and categorize the images into five groups based on the overlap between our pipeline’s predicted labels and the ReaL labels. We then conducted a detailed human evaluation on $250$ randomly sampled images ($50$ from each category). The reviewer is a PhD student familiar with ImageNet categories, using external references for fine-grained distinctions. The five categories are: (1) ReaL provides no label for the image; (2) Exact label set match between our method and ReaL; (3) Our labels are a superset of ReaL (we predict additional labels beyond ReaL); (4) ReaL is a superset of ours; and (5) Partial overlap (each provides some unique labels). Below we summarize the findings for each category.

\noindent \textbf{ReaL Has No Label (6.3\%).} First, it’s important to understand why ReaL might assign no label to an image. In ReaL’s annotations, $3,163$ images (about $6.3\%$ of the validation set) were discarded as having no label. This can occur if none of the expert models' proposed labels met the confidence criteria – for example, the image may not contain any object from the $1000$ ImageNet classes, or it is too ambiguous (e.g. only a small part of an object is visible, making identification uncertain). Our review revealed that about $16\%$ of the sampled images in this category clearly had no valid label (the content genuinely falls outside ImageNet’s classes or is unrecognizable), and another $20\%$ were borderline/unsure cases where a label could not be confidently assigned (as illustrated in Fig.~\ref{fig:qualitative_appendix} (a)). 
These findings are consistent with prior studies that report a non-trivial rate of label errors in ImageNet~\cite{vasudevan2022does}.
Importantly, we found that $64\%$ of the images in this category did contain at least one object from the ImageNet classes—indicating that ReaL either overlooked them due to its conservative filtering or excluded them due to annotation errors. Among these, our pipeline correctly recovered one or more missing labels in $94\%$ of the cases. This shows that our automated approach can effectively address many of the “missing label” scenarios that ReaL leaves unannotated. However, in about $6\%$ of these cases, while a valid object was present, our method still failed—typically due to poor segmentation (e.g., small or occluded objects), which led to missed predictions. In future work, we aim to explore strategies such as additional segmentation priors, multi-scale processing, and higher-resolution features to better handle these challenging cases.

\noindent \textbf{Exact Label Set Match (62.9\%).} In this category, our pipeline produced an \textit{exact match} with the ReaL multi-label annotations. This outcome indicates a strong agreement between our automatic method and the human multilabel annotation for those images. We further examined whether our method’s explanations (in the form of object masks) align with the predicted labels. Reassuringly, for 94\% of these images, the predicted mask(s) correspond closely to the actual object(s) of the given class, demonstrating that our model is looking at the correct regions when making the predictions. Fig.~\ref{fig:qualitative_appendix} (b) shows an example where the mask precisely highlights the target object, supporting the chosen label. In a small fraction ($6\%$), however, the masks were found to be noisy or misaligned – for instance, highlighting unrelated patches or only part of the object.
These few cases suggest that the model occasionally relies on context or struggles to precisely localize the object, even when it predicts the correct label.
We hypothesize that this stems from slight overfitting or bias in the classification head: the object proposals included in its training are automatically filtered with ReLabel~\cite{relabel} which could include false positives (see Fig.~\ref{fig:diagram} (c)). 
In addition, the ViT’s global self-attention enables a masked patch to propagate or receive information from the entire image, which may contribute to the observed bleed-through.
Nonetheless, these cases turn to be low-confidence predictions, and our pipeline’s mask-based approach largely mitigates such issues (see Fig.~\ref{fig:diagram} (d)). 

\noindent \textbf{Our Labels Superset of ReaL (13.1\%).} In these cases, our pipeline predicted one or more additional labels that were not present in ReaL’s annotation for the image (while still predicting all the labels that ReaL did). The key question is whether our extra labels are indeed valid or are false positives. We found that $74\%$ of the additional labels proposed by our method were correct upon human inspection – i.e. there was genuinely an instance of that class in the image, which ReaL’s labels had missed (see Fig.~\ref{fig:qualitative_appendix} (c)). This suggests that ReaL, despite being more comprehensive than the original single-label ImageNet, still missed a non-trivial number of valid labels. One reason is the design of the ReaL annotation process: if all the models in their proposal set confidently agreed on a single label (usually the original ImageNet label), the image was not sent for multilabel human review. This likely caused many secondary objects to be overlooked. Indeed, previous analyses indicate that roughly $20$–$30\%$ of ImageNet validation images contain multiple objects or multiple plausible labels~\cite{vasudevan2022does}. 

A single-label model ensemble tends to pick only the dominant object, so those images could receive no additional labels in ReaL’s pipeline. Our method, by contrast, explicitly searched for multiple objects via segmentation and was able to identify many of those missing labels. On the other hand, $26\%$ of the extra labels from our pipeline in this category turned out to be incorrect upon review. The most common failure mode (about $18\%$ out of the $26\%$) was that our model identified a region and assigned a related but wrong class to it – often because the true object category was not actually among the 1000 ImageNet classes (for instance, we assign \texttt{teddy bear} to stuffed toys in the image; Fig.~\ref{fig:qualitative_appendix} (f)). In these cases, the pipeline might detect a genuine object, but labels it as the closest known category. Such mistake roots in the fixed label space. The rest of the errors were more minor: in $4\%$ of cases, the mask was imperfect (covering only part of the object or blending objects) which led to a misclassification, and in another $4\%$ the mask was fine but the classifier simply made the wrong prediction for that region. Overall, however, the high percentage of correct new labels indicates that our pipeline can substantially augment ReaL by recalling additional valid labels that are missed.

\input{tables/qualitative}

\noindent \textbf{ReaL Superset of Ours (7.7\%).} This is the opposite scenario – here ReaL provided one or more labels that our pipeline failed to predict. The first point to note is whether ReaL’s extra labels are truly correct. We found that in about $16\%$ of such cases, the additional label from ReaL appears to be incorrect or at least very debatable. In ReaL’s annotation protocol, the threshold for including a label was set to ensure high recall. This means ReaL sometimes included labels for ambiguous instances to avoid missing a possible object, even if that label might not be clearly evident. For example, an object could be labeled as two different but visually similar classes if the annotators weren’t sure which it is, or a part of an object might get a separate label (e.g. labeling both “airplane” and “wing” in an image of an airplane). Some of these extra labels turn out, on closer examination, to be unnecessary or erroneous – they were essentially false positives introduced by an overly generous labeling policy. Aside from those, the remaining $84\%$ of ReaL’s additional labels were correct, meaning our pipeline genuinely missed detecting those objects or classes. We analyzed why our method missed these labels, and found a few recurring issues:
\begin{itemize}
    \item \textbf{Ambiguity in class definitions (56\%)} – In a majority of these cases, the image had an object that could plausibly belong to multiple closely related classes, or the ImageNet classes themselves overlap in scope. ReaL often handled this by assigning multiple labels to cover all bases, whereas our pipeline typically picked just one. For instance, an image of a certain dog breed might have been given two breed labels in ReaL because it was hard to distinguish (both labels were considered valid) – our model might only predict one of them. Similarly, images with objects that fall into an “X and part-of-X” situation (“car” and “car wheel”) or singular/plural duplicates (“sunglasses” vs “sunglass”) can legitimately have multiple labels (as shown in Fig.~\ref{fig:qualitative_appendix} (d, f)). Our pipeline, due to its single-instance mask proposal, might only tag the larger object or the more obvious class. The inherent ambiguity or overlapping taxonomy of ImageNet classes led to our method missing some labels that ReaL included. Notably, recent work has argued for collapsing such overlapping classes to improve evaluation~\cite{vasudevan2022does}.
    \item \textbf{Segmentation limitations (18\%)} – In these cases, our pipeline did not produce a separate mask for an object of interest, often because the object was very small, occluded, or touching another object. For example, if two objects were adjacent, MaskCut might generate one combined mask covering both, causing the model to predict only one class for that region (Fig.~\ref{fig:qualitative_appendix} (f)). Consequently, a valid label went missing simply because the object was not isolated by our proposals.
    \item \textbf{Low confidence pruning (6\%)} – Here, our model actually did predict the correct label, but with a confidence below our threshold $\tau=0.5$, so we filtered it out. In other words, the label was on our initial list but got discarded for not being confident enough. 
    \item \textbf{Misclassification (4\%)} – In the remaining small portion, the pipeline did propose a mask for the object and should have been able to predict the label, but it simply predicted the wrong class for that region. These are straightforward errors of the classification head on clear objects (e.g. mistaking one breed for another).
\end{itemize}

\noindent \textbf{Partial Overlap (9.9\%).} This category includes images where our pipeline and ReaL each identified some correct labels that the other missed. In other words, the two label sets intersect but neither is a strict subset of the other.  Our analysis found a mix of outcomes here: in $36\%$ of these cases, both our labels and ReaL’s labels were correct in what they included, but each method was incomplete – together they gave a more complete description of the image than either alone. This underscores how challenging it is to get a perfectly comprehensive label set, even with human annotators, and shows that our pipeline can complement the human labels by catching different subsets of objects. In $28\%$ of partial-overlap cases, ReaL’s unique labels were correct while our unique label was incorrect (so ReaL had the better coverage), whereas in $20\%$ of the cases it was the opposite – our extra label was correct and one of ReaL’s labels was actually incorrect. Finally, $16\%$ of the cases had errors on both sides (each provided at least one wrong label that the other did not). The reasons for these misses and mistakes mirror the patterns discussed above. Ambiguities in the image or label definitions often led to one method assigning an extra label that the other omitted; for example, ReaL might include an arguably present object that our pipeline’s mask proposals overlooked, or conversely our model might flag an object with a label that ReaL’s annotators were overly conservative about. Likewise, some of ReaL’s labels in this category turned out to be over-generalizations or slight mistakes, while some objects our model missed were due to segmentation or confidence issues as described. 

This qualitative evaluation shows that our automated relabeling pipeline aligns closely with the human-curated ReaL labels while offering meaningful improvements in several areas. In the majority of cases, our method either agrees with ReaL or correctly recovers additional labels that ReaL misses, demonstrating strong recall of true object classes. In particular, our pipeline successfully resolves many “no-label” cases and augments single-label annotations with valid secondary objects—addressing gaps in ReaL’s coverage due to its high-precision filtering protocol. To quantify this alignment, Table~\ref{tab:qualitative} presents an estimated breakdown of label agreement and disagreement categories, based on our human evaluation of $250$ sampled images.

At the same time, our method is not without limitations. It cannot fully resolve inherent ambiguities in the ImageNet taxonomy—such as overlapping, fine-grained, or under-specified categories—which can challenge both automated models and human annotators. Occasional errors also arise from imperfect segmentation, conservative thresholding, or incorrect predictions for ambiguous regions. However, these failure modes are relatively rare and often complementary to those found in ReaL. Overall, our multi-label pipeline provides a robust, interpretable, and scalable alternative to manual annotation, and can help identify missing or questionable labels. In practice, combining model-driven relabeling with human verification offers a promising path toward improving the quality of large-scale datasets like ImageNet.

\input{tables/ambiguous_class_pair}
\input{tables/fix_ambiguous}

\section{Handling Ambiguous Classes}
\label{app:fix_ambiguous_class}
Our method assumes one label per region, which is generally appropriate for datasets like COCO~\cite{lin2014microsoft} or VOC~\cite{everingham2010pascal}, but can break down under ImageNet’s taxonomy (see Fig.~\ref{fig:qualitative_appendix} (d,f)). Ambiguities arise in cases involving synonyms (e.g., \texttt{sunglass} vs. \texttt{sunglasses}), part–whole relationships (e.g., \texttt{airplane} vs. \texttt{wing}), or hierarchical categories (e.g., \texttt{wool} vs. \texttt{cardigan}). These lead to under-labeling when only one class in a semantically overlapping pair is assigned. We explore ways to mitigate this issue through targeted post-processing.

\subsection{Identifying Ambiguous Class Pairs}
To systematically identify such cases, we analyze the top $1000$ most frequently co-occurring class pairs in the ReaL~\cite{real} validation set (frequency $\geq 3$). We manually examine each pair and verify the possible ones that contain ambiguity by sampling $50$ images per class for visual comparison. This process yields $34$ ambiguous class pairs, listed in Table~\ref{tab:ambiguous_pairs}.

\subsection{Post-Processing with Co-occurrence Priors}
We propose two complementary post-processing strategies that leverage empirical co-occurrence statistics from ReaL and model-predicted soft labels to resolve ambiguity and enrich label completeness.

\noindent \textbf{Co-occurrence Prior Propagation.} We construct a symmetric co-occurrence matrix $C \in [0,1]^{K \times K}$ from ReaL, where $K=1000$ and $C_{ij}$ denotes the normalized frequency that class $j$ appears alongside class $i$ among the identified ambiguous pairs. For a predicted soft label vector $\mathbf{p} \in [0,1]^K$, we apply:
\begin{equation}
\mathbf{p}' = \operatorname{clip}(C \cdot \mathbf{p}, 0, 1),
\end{equation}
where $\mathbf{p}'$ is the adjusted soft label vector and clip denotes element-wise clamping to $[0,1]$. This propagates confidence from one class to its frequent ambiguous companion.

\noindent \textbf{Asymmetric Thresholding.} For each ambiguous class pair $(a, b)$, we derive class-specific thresholds $\tau_a$ and $\tau_b$ based on their conditional co-occurrence in ReaL. If class $a$ is the top-1 predicted label and the model's predicted confidence for class $b$ exceeds $\tau_b$, we add $b$ as an additional label by assigning it the same confidence score as $a$. This rule is applied symmetrically: $a$ is also added if $b$ is top-1 and $a$ exceeds $\tau_a$. These thresholds are designed to match the observed conditional probabilities $P(b \mid a)$ and $P(a \mid b)$, ensuring that added labels reflect the typical co-occurrence patterns in the data.

Let:
\begin{itemize}
    \item $N_a$, $N_b$ = number of images labeled with class $a$ and $b$ respectively,
    \item $N_{ab}$ = number of images labeled with both $a$ and $b$,
    \item $M_a$, $M_b$ = number of single-labeled images (with only $a$ or only $b$) to be upgraded with the missing class.
\end{itemize}

We estimate $M_a$, $M_b$ by solving:
\begin{equation}
P(b \mid a) = \frac{N_{ab} + M_b}{N_a + M_b}, \qquad
P(a \mid b) = \frac{N_{ab} + M_a}{N_b + M_a},
\end{equation}
which gives a closed-form solution for $M_a$ and $M_b$. We then cap the number of ambiguous instances added per class to ensure they do not exceed the number of images that contain class $a$ or $b$ individually (excluding co-occurrence):
\begin{align*}
M_a= \texttt{Max}(0, \texttt{Min}(M_a, N_a-N_{ab})), \\
M_b= \texttt{Max}(0, \texttt{Min}(M_b, N_b-N_{ab})).
\end{align*}
We rank single-labeled images by their softmax score on the missing class and select the top $M_a$ and $M_b$ examples accordingly. The lowest selected softmax score becomes the threshold $\tau_b$ (and similarly for $\tau_a$). These thresholds are derived based on the validation set and directly applied to the training set.
\input{tables/cross_arch_size}

\subsection{Training with Adjusted Labels}
We retrain models using our ambiguity-adjusted multi-label annotations, following the setup detailed in Supp.~\ref{app:cross_arch_size_exp_config}. Results are presented in Table~\ref{tab:fix_ambiguous}. Both correction strategies—Asymmetric Thresholding and Co-occurrence Prior Propagation—consistently improve over the original multi-label supervision across all architectures, input resolutions, and evaluation sets. For example, for ResNet-50, Asymmetric Thresholding improves top-1 accuracy by $+0.27$ (ReaL) and $+0.32$ (INv2), and mAP by $+0.35$ (ReaL) and $+0.68$ (INv2-ML). Co-occurrence Prior also yields performance gains, with up to $+0.30$ (ReaL, ResNet-101) and $+0.46$ (INv2, ViT-B/224) in top-1 accuracy, and $+0.46$ (ReaL, ResNet-101) and $+0.68$ (INv2-ML, ResNet-101) in mAP. 
These results validate the effectiveness of our proposed remedies and suggest that class-level ambiguity, which is rooted in the structure of the ImageNet taxonomy, can be effectively mitigated using priors derived from a lightweight calibration set (e.g., ReaL, whose size is roughly $3.65\%$ relative to the ImageNet training set).

\section{Additional Analyses}
\label{app:additional_exp}
\subsection{Robustness to Input Resolution (384×384)}
\label{app:cross_arch_size_full}
We extend the analysis in Sec.~\ref{sec:arch_size_transfer} to higher input resolution ($384^2$) for ViT models, following the DeiT3 training setup. Table~\ref{tab:cross_arch_size} presents results for training or fine-tuning with multi-label supervision at both $224^2$ and $384^2$ input resolutions.
Overall, the improvements from multi-label training persist across input resolutions. For fine-tuning at $384^2$ resolution, the largest gains on ReaL and INv2-ML w.r.t. top-1 accuracy are $+0.79$ (ViT-Base) and $+1.45$ (ViT-Small). And, in terms of mAP, the improvements are up to $+0.48$ and $+2.55$ on ReaL and INv2-ML, respectively.
End-to-end training also shows consistent trends: ViT-Small, ViT-Base, and ViT-Large yield gains of $+1.75$, $+0.48$, and $+0.12$ respectively in top-1 accuracy on INv2, and $+2.68$, $+1.44$, and $+0.85$ respectively in mAP on INv2-ML.
Under transfer learning, ViT-Large benefits notably. Fine-tuning improves COCO and VOC mAP by $+0.45$ and $+0.80$ respectively, while end-to-end multi-label training yields even larger boosts of $+2.19$ and $+1.96$.
These findings confirm the robustness and scalability of our multi-label supervision across model sizes and input resolutions.

\input{tables/koleo}

\input{tables/soft_label_training}
\subsection{Feature Diversity via k-NN Koleo Entropy}
\label{app:koleo}
We hypothesize that multi-label training encourages less representation collapse. 
Prior works~\cite{harun2025controlling, harun2024variables} show that higher entropy correlates with lower neural collapse and vice-versa, suggesting that representations with higher entropy transfer better to out-of-distribution datasets.
To evaluate transferability, following~\cite{harun2025controlling}, we compute $k$-nearest-neighbor ($k$-NN)-based entropy ($k=3$) for ImageNet, VOC, and COCO datasets. Results are summarized in Table~\ref{tab:koleo}. 
Across both ResNet and ViT, multi-label training consistently results in higher entropy than single-label supervision, supporting our claim that it leads to more diverse and transferable representations.

\subsection{Training with Soft Labels}
ReLabel~\cite{relabel} generates a $15{\times}15$ spatial label map using a pretrained classifier without the global pooling layer. During training, a random crop is applied to both the input image and the corresponding region in the label map. The soft label for that crop is then obtained by applying ROI-align on the label map using the crop coordinates, and normalized to sum to $1$ as the supervision target. We replicate this setup but replace ReLabel’s ImageNet-21K–pretrained teacher network with our own classification head trained on localized region crops using only ImageNet-1K supervision. As shown in Table~\ref{tab:soft_label}, our spatial label map achieves comparable or better classification accuracy—despite using less labeled data. Both methods outperform standard single-label training by $+0.7$–$0.9$ top-1 accuracy on IN-Val and ReaL, and our model converges faster and yields stronger performance at $100$ epochs, suggesting that our spatial labels are better aligned with object regions.

\input{tables/filter_maskcut}
\subsection{Filtering CutLER Masks with Our Labeler}
CutLER~\cite{wang2023cut} trains an unsupervised segmentation model using MaskCut-generated object proposals from ImageNet as initial pseudo-ground-truth masks. In its first iteration, all proposals—including low-quality or background masks—are retained because there is no mechanism to assess mask reliability. A segmentation model is then trained on these raw masks and later used to refine them in subsequent iterations. To isolate the effect of proposal quality, we focus exclusively on this first-iteration training stage.
Before training, we apply our region-level labeler to each released CutLER mask and compute its top-1 softmax confidence. Masks with confidence below a threshold of $0.5$ are discarded, removing roughly $12\%$ of proposals. We then train a Cascade Mask R-CNN~\cite{cai2018cascade} using either the original CutLER pseudo-masks or our filtered pseudo-masks (all derived from ImageNet). Evaluation is conducted on COCO in the standard zero-shot transfer setting. As shown in Table~\ref{tab:filter_maskcut}, filtering improves zero-shot segmentation accuracy, and even a more conservative threshold of $0.3$—removing only $6\%$ of masks—still provides a measurable gain. These results demonstrate that our region-level classifier can serve as an effective post-hoc filter for noisy pseudo-masks—enhancing the quality of the supervision signal and potentially improving downstream segmentation performance even before iterative refinement.

\input{figures/annotation_tool}
\subsection{Interactive Annotation Tool}
Our localized labeler can be repurposed for interactive region-level annotation. Given a user-specified bounding box, we apply ROI-Align to extract pooled features from the ViT backbone and pass them to our labeler head, which outputs a ranked list of ImageNet class predictions with confidence scores. This enables fast and flexible labeling of arbitrary regions. As illustrated in Fig.~\ref{fig:annotation_tool}, our region-level labeler successfully predicts labels for multiple objects in the scene, even when the original and ReaL labels only include a single class.
Notably, even when the target object is not explicitly represented in the ImageNet label space (e.g., a hiking boot), the labeler tends to predict semantically related classes (e.g., running shoe) with appropriately lower confidence. This behavior supports broad-category tagging and reduces the burden of exhaustive class coverage. This tool can accelerate human annotation for downstream datasets or for refining labels in existing benchmarks like ImageNet.

%% file: figures/sam_example.tex
\begin{figure}[t]
    \centering
\includegraphics[width=0.49\textwidth]{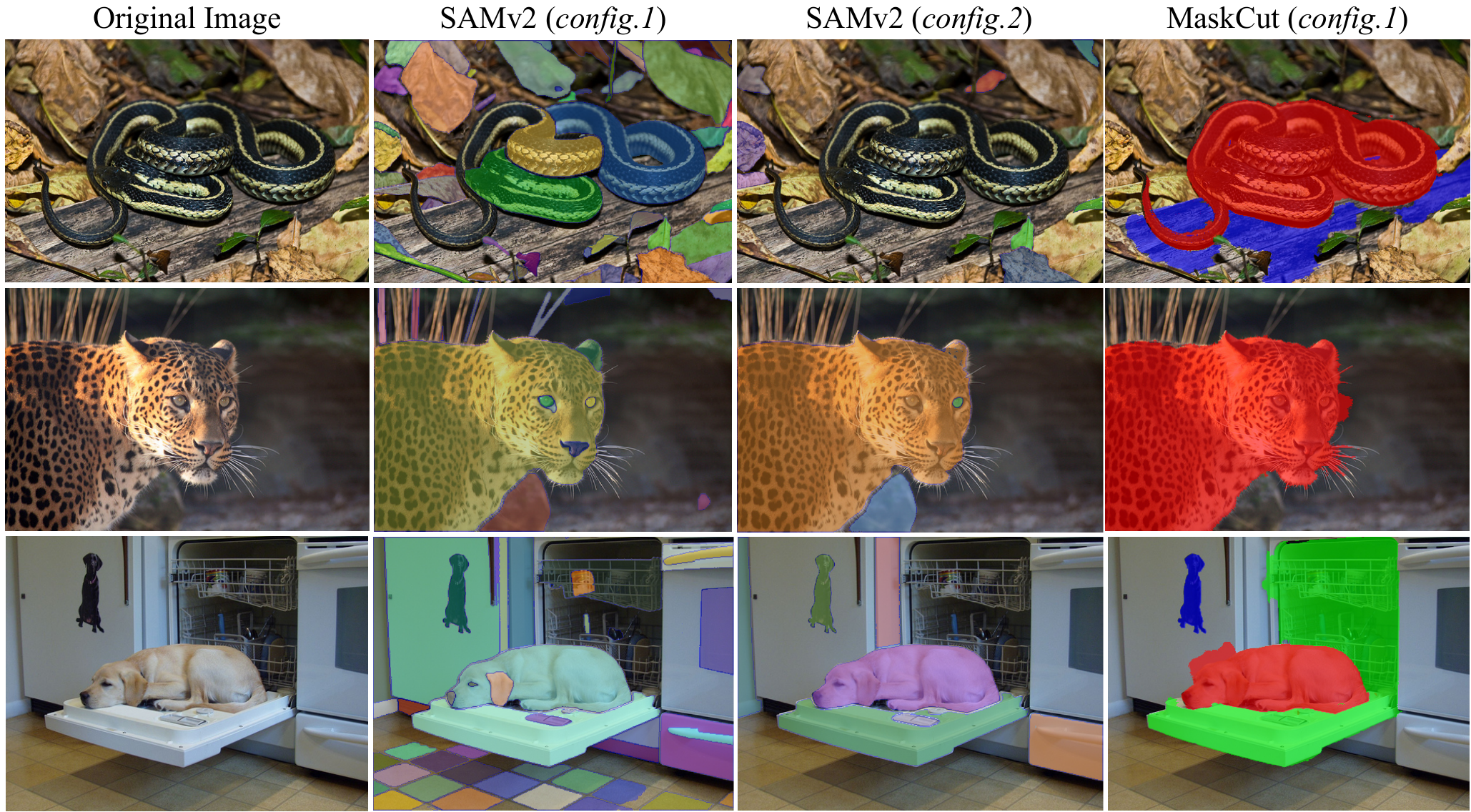}
    \caption{Comparison of mask proposals from SAMv2~\cite{ravi2024sam} (two configurations) and MaskCut~\cite{wang2023cut}. While SAM can produce fine-grained masks under certain settings, it often over-segments or misses key objects depending on the image, highlighting its sensitivity to hyperparameters. In contrast, MaskCut generates more consistent, instance-level masks across diverse images using a fixed configuration, making it more suitable for large-scale object proposal generation.}
    \label{fig:sam}
\end{figure}

%% file: tables/train_statistics.tex
\begin{table}[t]
\caption{
Distribution of the number of unique labels ($\mathbf{k}$) predicted per image in the relabeled ImageNet training set, using a softmax confidence threshold of $\tau=0.5$. The last column (Avg.) reports the average number of labels per image across the full training set.
}
\label{tab:train_statistics}
\centering
\resizebox{0.4\textwidth}{!}{
\begin{tabular}{ccccc|c}
\hline
$\mathbf{k=0}$ & $\mathbf{k=1}$ & $\mathbf{k=2}$ & $\mathbf{k=3}$ & $\mathbf{k\ge4}$ & \textbf{Avg.} \\ \hline
0.59\%      & 78.9\%      & 17.8\%      & 2.5\%       & 0.28\%                    & 1.23          \\ \hline
\end{tabular}
}
\end{table}

%% file: tables/sweep_label_setup.tex
\begin{table*}[t]
  \caption{
  Evaluation of different relabeling configurations on multi-label benchmarks. We sweep thresholding strategies (probability vs. fixed threshold) and choice of global supervision (original vs. predicted) using a ResNet-50 trained on ImageNet for 100 epochs. The best and second best performance are highlighted by bold and underline, respectively. 
  }
  \label{tab:sweep_label_setup}
\centering
\resizebox{0.8\textwidth}{!}{
\begin{tabular}{l l | c c c c c | c c c}
\toprule
\multirow{2}{*}{\textbf{\begin{tabular}[c]{@{}c@{}}Local \\ Label\end{tabular}}} & \multirow{2}{*}{\textbf{\begin{tabular}[c]{@{}c@{}}Global \\ Label\end{tabular}}} & 
\multicolumn{5}{c|}{\textbf{Top-1 Acc} $\uparrow$} &
\multicolumn{3}{c}{\textbf{mAP} $\uparrow$} \\
\cmidrule{3-10}
\multicolumn{2}{c|}{} &
\textbf{IN-Val} & 
\textbf{ReaL} & 
\textbf{IN-Seg} & 
\textbf{IN-v2} & 
\textbf{INv2-ML} &
\textbf{ReaL} & 
\textbf{IN-Seg} & 
\textbf{INv2-ML} \\
\midrule
\multicolumn{2}{c|}{\textit{Original Label}}& 77.6 & 84.1 & 84.1 & 65.7 & 77.7 & 87.1 & 87.6 & 72.7  \\
\midrule
Hard ($\tau=0.5$)& \textit{None}& 77.3& 84.6& 85.0& 66.0& 79.2& 86.9& 88.1& 73.4\\
Soft & \textit{None}& 77.4& 84.3& 84.5& 66.2& 78.4& 87.2& 88.1& 73.0\\
Hard ($\tau=0.8$) & Pred & 77.7 & \underline{84.6} & 84.9 & 66.1 & 78.7 & 87.0 & 88.0 & 72.3 \\
Soft  & Pred & 77.6 & 84.5 & 85.0 & 66.5 & \underline{79.3} & \underline{87.6} & 88.2 & \underline{74.5} \\
Hard ($\tau=0.9$) & Original  & \textbf{78.0} & 84.5 & \underline{85.0} & \underline{66.6} & 79.0 & 87.6 & \underline{88.3} & 74.2 \\
Soft   & Original  & \underline{77.8} & \textbf{84.7} & \textbf{85.1} & \textbf{66.8} & \textbf{79.6} & \textbf{87.7} & \textbf{88.5} & \textbf{74.7} \\
\bottomrule
\end{tabular}
}
\end{table*}

%% file: figures/qualitative_appendix.tex
\begin{figure*}[t]
    \centering
    \includegraphics[width=0.99\textwidth]{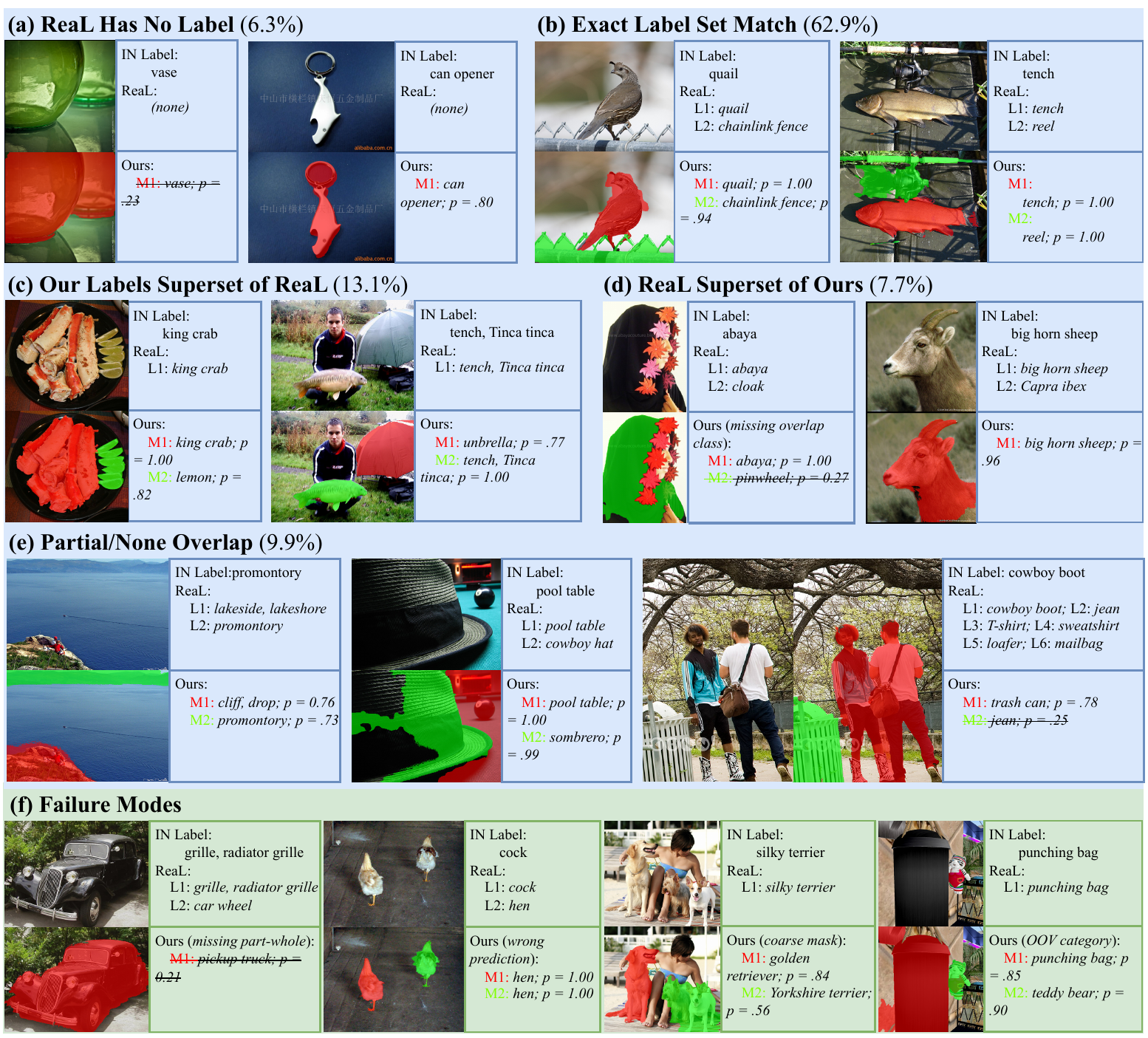}
    \vspace{-3mm}
    \caption{Additional qualitative comparisons between our multi-label annotations and those from ImageNet and ReaL~\cite{real}.
Blue panels (a-e): Examples categorized by the degree of label overlap with ReaL. Labels pruned due to low confidence are indicated by strikethrough (\sout{deleted}).
Green panels (f): Common failure modes, including missed part–whole relations, fine-grained confusion, and proposals outside the label space (out-of-vocabulary, OOV). 
}
    \label{fig:qualitative_appendix}
\end{figure*}

%% file: tables/qualitative.tex
\setlength{\fboxsep}{1pt} 
\begin{table}[t]
\vspace{-1mm}
\caption{Human evaluation breakdown of agreement between our relabeling and ReaL~\cite{real} on the ImageNet validation set. Based on detailed review of 250 sampled images, we estimate the proportion of images falling into three categories: (a) ReaL provides no label (6.33\%), (b) ReaL’s label set exactly matches the human-determined ground truth (estimated 75.6\%), and (c) ReaL misses or mislabels one or more ground-truth classes (estimated 18.1\%). Each group is further subdivided by the quality of ReaL’s annotations and whether our method successfully recovers missing or incorrect labels. Percentages indicate estimated proportions over the full 50,000 ImageNet validation set. \colorbox[HTML]{FEE796}{Orange} cells denote ReaL’s labeling status and annotation breakdown; \colorbox[HTML]{E3F2D9}{green} cells indicate our method’s fixes or misses. GT refers to our human-verified ground truth.
}
\label{tab:qualitative}
\centering
\scriptsize
\begin{adjustbox}{width=0.49\textwidth}
\begin{tabular}{ccccc}
\hline
\rowcolor[HTML]{F2F2F2} 
\multicolumn{5}{c}{\cellcolor[HTML]{FEE796} (a) ReaL Has No Label: 6.33\%} \\ \hline
\rowcolor[HTML]{FEE796} 
\multicolumn{1}{c|}{\cellcolor[HTML]{F2F2F2}} &
  \multicolumn{2}{c|}{\cellcolor[HTML]{FEE796}Clear No Label: 1.01\%} &
  \multicolumn{2}{c}{\cellcolor[HTML]{FEE796}Has Label: 4.05\%} \\ \cline{2-5} 
\multicolumn{1}{c|}{\multirow{-2}{*}{\cellcolor[HTML]{F2F2F2}\begin{tabular}[c]{@{}c@{}}\textit{Unsure}: \\ \textit{1.27\%}\end{tabular}}} &
  \multicolumn{1}{c|}{\cellcolor[HTML]{E3F2D9}\begin{tabular}[c]{@{}c@{}}Ours No Label: \\ 0.51\%\end{tabular}} &
  \multicolumn{1}{c|}{\cellcolor[HTML]{E3F2D9}\begin{tabular}[c]{@{}c@{}}Ours Has Label: \\ 0.51\%\end{tabular}} &
  \multicolumn{1}{c|}{\cellcolor[HTML]{E3F2D9}\begin{tabular}[c]{@{}c@{}}Ours Fix: \\ 3.80\%\end{tabular}} &
  \multicolumn{1}{c}{\cellcolor[HTML]{E3F2D9}\begin{tabular}[c]{@{}c@{}}Not Fixed: \\ 0.25\%\end{tabular}} \\ \hline
\end{tabular}
\end{adjustbox}
\vspace{1mm}

\begin{adjustbox}{width=0.49\textwidth}
\begin{tabular}{
>{\columncolor[HTML]{E3F2D9}}c 
>{\columncolor[HTML]{E3F2D9}}c 
>{\columncolor[HTML]{E3F2D9}}c 
>{\columncolor[HTML]{E3F2D9}}c 
>{\columncolor[HTML]{E3F2D9}}c 
>{\columncolor[HTML]{E3F2D9}}c }
\hline
\multicolumn{6}{c}{\cellcolor[HTML]{FEE796}(b) ReaL Exact Match to GT: 75.6\%} \\ \hline
\multicolumn{2}{c|}{\cellcolor[HTML]{E3F2D9}Ours Match: 62.9\%} &
  \multicolumn{2}{c|}{\cellcolor[HTML]{E3F2D9}Ours Missed: 6.47\%} &
  \multicolumn{2}{c}{\cellcolor[HTML]{E3F2D9}Ours Wrong: 6.18\%} \\ \hline
\multicolumn{1}{c|}{\cellcolor[HTML]{E3F2D9}Good Mask:} &
  \multicolumn{1}{c|}{\cellcolor[HTML]{E3F2D9}Noisy Mask:} &
  \multicolumn{1}{c|}{\cellcolor[HTML]{E3F2D9}Ambig. Class:} &
  \multicolumn{1}{c|}{\cellcolor[HTML]{E3F2D9}Other:} &
  \multicolumn{1}{c|}{\cellcolor[HTML]{E3F2D9}Due to Mask:} &
  Wrong Pred.: \\
\multicolumn{1}{c|}{\cellcolor[HTML]{E3F2D9}59.1\%} &
  \multicolumn{1}{c|}{\cellcolor[HTML]{E3F2D9}3.77\%} &
  \multicolumn{1}{c|}{\cellcolor[HTML]{E3F2D9}4.31\%} &
  \multicolumn{1}{c|}{\cellcolor[HTML]{E3F2D9}2.16\%} &
  \multicolumn{1}{c|}{\cellcolor[HTML]{E3F2D9}3.43\%} &
  2.75\% \\ \hline
\end{tabular}
\end{adjustbox}
\vspace{1mm}

\begin{adjustbox}{width=0.49\textwidth}
\begin{tabular}{cccccc}
\hline
\rowcolor[HTML]{FEE796} 
\multicolumn{6}{c}{\cellcolor[HTML]{FEE796} (c) ReaL Not Match to GT: 18.1\%} \\ \hline
\rowcolor[HTML]{FEE796} 
\multicolumn{2}{c|}{\cellcolor[HTML]{FEE796}Missing Label: 12.3\%} &
  \multicolumn{2}{c|}{\cellcolor[HTML]{FEE796}Error Label: 2.69\%} &
  \multicolumn{2}{c}{\cellcolor[HTML]{FEE796}Missing and Error: 3.13\%} \\ \hline
\rowcolor[HTML]{E3F2D9} 
\multicolumn{1}{c|}{\cellcolor[HTML]{E3F2D9}Ours Fix:} &
  \multicolumn{1}{c|}{\cellcolor[HTML]{E3F2D9}Not Fixed:} &
  \multicolumn{1}{c|}{\cellcolor[HTML]{E3F2D9}Ours Fix:} &
  \multicolumn{1}{c|}{\cellcolor[HTML]{E3F2D9}Not Fixed:} &
  \multicolumn{1}{c|}{\cellcolor[HTML]{E3F2D9}Ours Fix:} &
  Not Fixed: \\
\rowcolor[HTML]{E3F2D9} 
\multicolumn{1}{c|}{\cellcolor[HTML]{E3F2D9}10.5\%} &
  \multicolumn{1}{c|}{\cellcolor[HTML]{E3F2D9}1.79\%} &
  \multicolumn{1}{c|}{\cellcolor[HTML]{E3F2D9}1.79\%} &
  \multicolumn{1}{c|}{\cellcolor[HTML]{E3F2D9}0.90\%} &
  \multicolumn{1}{c|}{\cellcolor[HTML]{E3F2D9}2.24\%} &
  0.90\% \\ \hline
\end{tabular}
\end{adjustbox}
\vspace{-5mm}
\end{table}

%% file: tables/ambiguous_class_pair.tex
\begin{table*}[t]
\caption{List of identified ambiguous class pairs in ImageNet. For each pair, we report class-wise occurrence counts, as well as conditional label confidence: $\text{Conf}(A|B)$ is the proportion of times class A appears given B is present, and vice versa.}
\label{tab:ambiguous_pairs}
\centering
\resizebox{0.95\textwidth}{!}{
\begin{tabular}{cllcccc}
\hline
\multicolumn{1}{c}{\textbf{Co-occurrence}} &
  \multicolumn{1}{c}{\textbf{Class $\mathbf{A}$}} &
  \multicolumn{1}{c}{\textbf{Class $\mathbf{B}$}} &
  \textbf{Freq($\mathbf{A}$)} &
  \textbf{Freq($\mathbf{B}$)} &
  \textbf{Conf($\mathbf{A|B}$)} &
  \textbf{Conf($\mathbf{B|A}$)} \\ \hline
153 & sunglass                        & sunglasses, dark glasses, shades & 175 & 168 & 0.91 & 0.87 \\
135 & laptop, laptop computer         & notebook, notebook computer      & 157 & 151 & 0.89 & 0.86 \\
88  & monitor                         & screen, CRT screen               & 227 & 101 & 0.87 & 0.39 \\
87  & maillot                         & maillot, tank suit               & 106 & 119 & 0.73 & 0.82 \\
84  & car wheel                         & grille, radiator grille              & 232  & 143  & 0.58 & 0.36 \\
55  & missile                         & projectile, missile              & 70  & 66  & 0.83 & 0.79 \\
51  & suit, suit of clothes           & Windsor tie                      & 126 & 83  & 0.61 & 0.40 \\
45  & space bar                       & typewriter keyboard              & 107 & 70  & 0.64 & 0.42 \\
35  & car wheel                       & convertible                      & 232 & 61  & 0.57 & 0.15 \\
34  & bookcase                        & bookshop, bookstore, bookstall   & 116 & 77  & 0.44 & 0.29 \\
33  & bathtub, bathing tub, bath, tub & tub, vat                         & 87  & 42  & 0.79 & 0.38 \\
30  & airliner                        & wing                             & 79  & 84  & 0.36 & 0.38 \\
26  & cassette                        & cassette player                  & 69  & 62  & 0.42 & 0.38 \\
23  & warplane, military plane        & wing                             & 57  & 84  & 0.27 & 0.40 \\
20  & ox                              & oxcart                           & 47  & 65  & 0.31 & 0.43 \\
18  & analog clock                    & wall clock                       & 48  & 86  & 0.21 & 0.38 \\
17  & car wheel                       & pickup, pickup truck             & 232 & 54  & 0.31 & 0.07 \\
17  & convertible                       & grille, radiator grille             & 61 & 143  & 0.12 & 0.28 \\
15  & grille, radiator grille                       & pickup, pickup truck             & 143 & 54  & 0.28 & 0.10 \\
14  & assault rifle, assault gun      & rifle                            & 85  & 58  & 0.24 & 0.16 \\
14  & car wheel                       & minivan                          & 232 & 54  & 0.26 & 0.06 \\
14  & car wheel                       & Model T                          & 232 & 60  & 0.23 & 0.06 \\
13  & grille, radiator grille                       & sports car, sport car                          & 143 & 45  & 0.29 & 0.10 \\
12  & beach wagon, station wagon       & grille, radiator grille                          & 51 & 143  & 0.08 & 0.23 \\
10  & drum, membranophone, tympan     & steel drum                       & 60  & 76  & 0.13 & 0.17 \\
10  & flagpole, flagstaff             & pole                             & 70  & 85  & 0.12 & 0.14 \\
7   & barrel, cask                    & rain barrel                      & 45  & 64  & 0.11 & 0.16 \\
7   & cardigan                        & wool, woolen, woollen            & 58  & 102 & 0.07 & 0.12 \\
7   & grille, radiator grille                        & minivan            & 143  & 54 & 0.13 & 0.05 \\
5   & abaya                     & cloak            & 55  & 51  & 0.10 & 0.09 \\
5   & grille, radiator grille                     & Model T            & 143  & 60  & 0.08 & 0.03 \\
4   & grille, radiator grille                     & racer, race car, racing car            & 143  & 48  & 0.08 & 0.03 \\
4   & rule, ruler                     & slide rule, slipstick            & 81  & 39  & 0.10 & 0.05 \\
3   & measuring cup                   & cup                              & 54  & 140 & 0.02 & 0.06 \\
3   & espresso maker                  & espresso                         & 41  & 57  & 0.05 & 0.07 \\
\hline
\end{tabular}
}
\end{table*}

%% file: tables/fix_ambiguous.tex
\begin{table*}[t]
\caption{Effect of ambiguity resolution strategies on ImageNet multi-label training. We evaluate two approaches for handling ambiguous class pairs: (1) Coexistence Prior, which adjusts label targets using a class co-occurrence matrix derived from ReaL; and (2) Threshold Pairing, which applies asymmetric confidence thresholds to resolve semantic overlaps. Both methods improve upon the baseline IN1k-Mul supervision across all evaluation sets. Best performances are highlighted in bold.}
\label{tab:fix_ambiguous}
\centering
\scriptsize
\begin{adjustbox}{width=0.95\textwidth}
\begin{tabular}{cclcccccccc}
\toprule
\multirow{2}{*}{\textbf{Model}} & \multirow{2}{*}{\textbf{Image}} & \multicolumn{1}{c}{\multirow{2}{*}{\textbf{Method}}} & \multicolumn{5}{c}{\textbf{Top-1 Acc $\uparrow$}} & \multicolumn{3}{c}{\textbf{Multi-Label: mAP $\uparrow$}} \\
\cmidrule(lr){4-8} \cmidrule(lr){9-11} 
 & & & \textbf{IN-Val} & \textbf{ReaL} & \textbf{IN-Seg} & \textbf{INv2} & \textbf{INv2-ML} & \textbf{ReaL} & \textbf{IN-Seg} & \textbf{INv2-ML} \\
\cmidrule{1-11}
\multirow{4}{*}{ResNet-50} &
  \multirow{4}{*}{$224^2$} &
  Single-label E2E &
  78.2 &
  84.1 &
  84.4 &
  66.1 &
  78.2 &
  87.0 &
  87.7 &
  72.3 \\
  \cmidrule{3-11}
 &
   &
  Multi-label E2E &
  78.7 &
  85.6 &
  85.5 &
  67.4 &
  81.0 &
  88.2 &
  88.8 &
  76.2 \\
 &
   &
  + Threshold Pairing &
  78.8 &
  \textbf{85.9} &
  \textbf{86.0} &
  \textbf{67.8} &
  \textbf{81.1} &
  88.5 &
  89.3 &
  \textbf{76.9} \\
 &
   &
  + Coexistence Prior &
  \textbf{78.9} &
  85.8 &
  85.9 &
  67.4 &
  80.9 &
  \textbf{88.5} &
  \textbf{89.3} &
  76.9 \\ \cmidrule{1-11}
\multirow{4}{*}{ResNet-101} &
  \multirow{4}{*}{$224^2$} &
  Single-label E2E &
  79.3 &
  84.7 &
  85.3 &
  68.2 &
  80.0 &
  87.1 &
  88.2 &
  72.7 \\
  \cmidrule{3-11}
 &
   &
  Multi-label E2E &
  80.2 &
  86.7 &
  86.9 &
  69.2 &
  82.3 &
  89.0 &
  89.9 &
  77.4 \\
 &
   &
  + Threshold Pairing &
  \textbf{80.6} &
  86.9 &
  87.0 &
  69.2 &
  \textbf{83.2} &
  89.3 &
  90.1 &
  \textbf{78.8} \\
 &
   &
  + Coexistence Prior &
  80.5 &
  \textbf{87.0} &
  \textbf{87.1} &
  \textbf{69.5} &
  83.0 &
  \textbf{89.5} &
  \textbf{90.2} &
  78.8 \\ \cmidrule{1-11}
\multirow{8}{*}{ViT-base} &
  \multirow{4}{*}{$224^2$} &
  Single-label E2E &
  \textbf{83.7} &
  88.2 &
  88.2 &
  73.6 &
  85.8 &
  90.6 &
  90.9 &
  80.3 \\
  \cmidrule{3-11}
 &
   &
  Multi-label E2E &
  83.4 &
  88.8 &
  89.1 &
  73.9 &
  86.8 &
  90.7 &
  91.5 &
  81.8 \\
 &
   &
  + Threshold Pairing &
  83.5 &
  88.8 &
  \textbf{89.1} &
  73.9 &
  \textbf{87.5} &
  90.9 &
  \textbf{91.7} &
  82.4 \\
 &
   &
  + Coexistence Prior &
  83.5 &
  \textbf{88.9} &
  88.9 &
  \textbf{74.4} &
  87.2 &
  \textbf{91.1} &
  91.6 &
  \textbf{82.7} \\ \cmidrule{2-11} 
 &
  \multirow{4}{*}{$384^2$} &
  Single-label E2E &
  \textbf{85.0} &
  88.8 &
  88.8 &
  74.7 &
  87.3 &
  91.0 &
  91.5 &
  81.5 \\
  \cmidrule{3-11}
 &
   &
  Multi-label E2E &
  84.6 &
  89.4 &
  89.5 &
  75.2 &
  88.0 &
  91.2 &
  91.9 &
  82.9 \\
 &
   &
  + Threshold Pairing &
  84.6 &
  89.6 &
  89.5 &
  75.3 &
  88.6 &
  91.4 &
  92.0 &
  83.4 \\
 &
   &
  + Coexistence Prior &
  84.6 &
  \textbf{89.7} &
  \textbf{89.6} &
  \textbf{75.4} &
  \textbf{88.6} &
  \textbf{91.5} &
  \textbf{92.1} &
  \textbf{83.8} \\ \hline
\end{tabular}
\end{adjustbox}
\end{table*}

%% file: tables/cross_arch_size.tex
\begin{table*}
\setlength{\arrayrulewidth}{0.6pt}
\arrayrulecolor{black} 
\caption{End-to-end training and transfer performance with our multi-label annotations. We compare single-label (Single-label E2E) training, fine-tuning with our multi-labels (+ Multi-label FT), and end-to-end multi-label training (Multi-label E2E) across various backbones and input sizes. Best results are highlighted in bold. Our multi-label supervision improves in-domain performance on ImageNet and its variants, and yields consistent gains in downstream multi-label transfer to COCO and VOC. 
}
\vspace{-1mm}
\label{tab:cross_arch_size}
\centering
\resizebox{0.99\textwidth}{!}{
\begin{tabular}{ccccccccccccc}
\toprule
\multicolumn{3}{c}{} & \multicolumn{8}{c}{\textbf{ImageNet Evaluation}} & \multicolumn{2}{c}{\textbf{Transfer Learning}} \\
\cmidrule(lr){4-11} \cmidrule(lr){12-13}
\textbf{Model} & \textbf{Image} & \textbf{Supervision} & \multicolumn{5}{c}{\textbf{Top-1 Acc $\uparrow$}} & \multicolumn{3}{c}{\textbf{Multi-Label: mAP $\uparrow$}} & \multicolumn{2}{c}{\textbf{mAP $\uparrow$}} \\
\cmidrule(lr){4-8} \cmidrule(lr){9-11} \cmidrule(lr){12-13}
 & & & \textbf{IN-Val} & \textbf{ReaL} & \textbf{IN-Seg} & \textbf{INv2} & \textbf{INv2-ML} & \textbf{ReaL} & \textbf{IN-Seg} & \textbf{INv2-ML} & \textbf{COCO} & \textbf{VOC} \\
\hline
\rowcolor[HTML]{F2F2F2} 
\cellcolor[HTML]{F2F2F2} &
  \cellcolor[HTML]{F2F2F2} &
  Single-label E2E &
  81.4 &
  87.0 &
  87.2 &
  70.7 &
  \multicolumn{1}{c}{\cellcolor[HTML]{F2F2F2}83.1} &
  89.0 &
  89.7 &
  75.6 &
  79.1 &
  91.0 \\
\rowcolor[HTML]{E3F2D9} 
\cellcolor[HTML]{F2F2F2} &
  \cellcolor[HTML]{F2F2F2} &
  +Multi-label FT &
  81.5 &
  87.7 &
  87.8 &
  70.8 &
  \multicolumn{1}{c}{\cellcolor[HTML]{E3F2D9}84.1} &
  90.1 &
  90.9 &
  80.2 &
  81.0 &
  91.9 \\
\rowcolor[HTML]{C9E4B4} 
\cellcolor[HTML]{F2F2F2} &
  \multirow{-3}{*}{\cellcolor[HTML]{F2F2F2}$224^2$} &
  Multi-label E2E &
  \textbf{82.0} &
  \textbf{88.1} &
  \textbf{88.3} &
  \textbf{72.2} &
  \multicolumn{1}{c}{\cellcolor[HTML]{C9E4B4}\textbf{85.1}} &
  \textbf{90.3} &
  \textbf{91.1} &
  \textbf{80.7} &
  \textbf{83.3} &
  \textbf{93.3} \\ \cline{3-13}
\rowcolor[HTML]{F2F2F2} 
\cellcolor[HTML]{F2F2F2} &
  \cellcolor[HTML]{F2F2F2} &
  Single-label E2E &
  83.2 &
  88.2 &
  88.2 &
  72.6 &
  \multicolumn{1}{c}{\cellcolor[HTML]{F2F2F2}85.1} &
  90.5 &
  91.0 &
  79.6 &
  84.5 &
  92.8 \\
\rowcolor[HTML]{E3F2D9} 
\cellcolor[HTML]{F2F2F2} &
  \cellcolor[HTML]{F2F2F2} &
  +Multi-label FT &
  83.3 &
  88.8 &
  89.0 &
  73.2 &
  \multicolumn{1}{c}{\cellcolor[HTML]{E3F2D9}86.6} &
  91.0 &
  91.7 &
  82.2 &
  85.4 &
  93.9 \\
\rowcolor[HTML]{C9E4B4} 
\multirow{-6}{*}{\cellcolor[HTML]{F2F2F2}ViT-small} &
  \multirow{-3}{*}{\cellcolor[HTML]{F2F2F2}$384^2$} &
  Multi-label E2E &
  \textbf{83.5} &
  \textbf{88.9} &
  \textbf{89.1} &
  \textbf{74.4} &
  \multicolumn{1}{c}{\cellcolor[HTML]{C9E4B4}\textbf{87.0}} &
  \textbf{91.0} &
  \textbf{91.8} &
  \textbf{82.3} &
  \textbf{87.3} &
  \textbf{94.9} \\ \hline
\rowcolor[HTML]{F2F2F2} 
\cellcolor[HTML]{F2F2F2} &
  \cellcolor[HTML]{F2F2F2} &
  Single-label E2E &
  \textbf{83.7} &
  88.2 &
  88.2 &
  73.6 &
  \multicolumn{1}{c}{\cellcolor[HTML]{F2F2F2}85.8} &
  90.6 &
  90.9 &
  80.3 &
  83.0 &
  92.7 \\
\rowcolor[HTML]{E3F2D9} 
\cellcolor[HTML]{F2F2F2} &
  \cellcolor[HTML]{F2F2F2} &
  +Multi-label FT &
  83.6 &
  \textbf{88.9} &
  88.8 &
  \textbf{74.0} &
  \multicolumn{1}{c}{\cellcolor[HTML]{E3F2D9}\textbf{86.9}} &
  90.6 &
  91.3 &
  81.3 &
  83.8 &
  93.6 \\
\rowcolor[HTML]{C9E4B4} 
\cellcolor[HTML]{F2F2F2} &
  \multirow{-3}{*}{\cellcolor[HTML]{F2F2F2}$224^2$} &
  Multi-label E2E &
  83.4 &
  88.8 &
  \textbf{89.1} &
  73.9 &
  \multicolumn{1}{c}{\cellcolor[HTML]{C9E4B4}86.8} &
  \textbf{90.7} &
  \textbf{91.5} &
  \textbf{81.8} &
  \textbf{84.7} &
  \textbf{94.5} \\ \cline{3-13} 
\rowcolor[HTML]{F2F2F2} 
\cellcolor[HTML]{F2F2F2} &
  \cellcolor[HTML]{F2F2F2} &
  Single-label E2E &
  \textbf{85.0} &
  88.8 &
  88.8 &
  74.7 &
  \multicolumn{1}{c}{\cellcolor[HTML]{F2F2F2}87.3} &
  91.0 &
  91.5 &
  81.5 &
  86.7 &
  94.3 \\
\rowcolor[HTML]{E3F2D9} 
\cellcolor[HTML]{F2F2F2} &
  \cellcolor[HTML]{F2F2F2} &
  +Multi-label FT &
  84.9 &
  \textbf{89.6} &
  \textbf{89.6} &
  75.1 &
  \multicolumn{1}{c}{\cellcolor[HTML]{E3F2D9}\textbf{88.1}} &
  \textbf{91.5} &
  \textbf{92.2} &
  \textbf{83.9} &
  87.3 &
  94.9 \\
\rowcolor[HTML]{C9E4B4} 
\multirow{-6}{*}{\cellcolor[HTML]{F2F2F2}ViT-base} &
  \multirow{-3}{*}{\cellcolor[HTML]{F2F2F2}$384^2$} &
  Multi-label E2E &
  84.6 &
  89.4 &
  89.5 &
  \textbf{75.2} &
  \multicolumn{1}{c}{\cellcolor[HTML]{C9E4B4}88.0} &
  91.2 &
  91.9 &
  82.9 &
  \textbf{87.7} &
  \textbf{95.3} \\ \hline
\rowcolor[HTML]{F2F2F2} 
\cellcolor[HTML]{F2F2F2} &
  \cellcolor[HTML]{F2F2F2} &
  Single-label E2E &
  84.6 &
  88.6 &
  88.5 &
  74.7 &
  \multicolumn{1}{c}{\cellcolor[HTML]{F2F2F2}87.1} &
  90.8 &
  91.2 &
  81.4 &
  84.8 &
  93.4 \\
\rowcolor[HTML]{E3F2D9} 
\cellcolor[HTML]{F2F2F2} &
  \cellcolor[HTML]{F2F2F2} &
  +Multi-label FT &
  \textbf{84.6} &
  \textbf{89.3} &
  \textbf{89.4} &
  75.1 &
  \multicolumn{1}{c}{\cellcolor[HTML]{E3F2D9}\textbf{88.2}} &
  \textbf{91.1} &
  91.9 &
  \textbf{83.3} &
  85.2 &
  94.4 \\
\rowcolor[HTML]{C9E4B4} 
\cellcolor[HTML]{F2F2F2} &
  \multirow{-3}{*}{\cellcolor[HTML]{F2F2F2}$224^2$} &
  Multi-label E2E &
  84.3 &
  89.3 &
  89.4 &
  \textbf{74.9} &
  \multicolumn{1}{c}{\cellcolor[HTML]{C9E4B4}87.9} &
  91.2 &
  \textbf{91.9} &
  83.0 &
  \textbf{86.4} &
  \textbf{95.0} \\ \cline{3-13} 
\rowcolor[HTML]{F2F2F2} 
\cellcolor[HTML]{F2F2F2} &
  \cellcolor[HTML]{F2F2F2} &
  Single-label E2E &
  \textbf{85.9} &
  89.8 &
  89.7 &
  76.3 &
  \multicolumn{1}{c}{\cellcolor[HTML]{F2F2F2}89.1} &
  91.6 &
  92.1 &
  83.1 &
  87.3 &
  94.2 \\
\rowcolor[HTML]{E3F2D9} 
\cellcolor[HTML]{F2F2F2} &
  \cellcolor[HTML]{F2F2F2} &
  +Multi-label FT &
  85.5 &
  \textbf{90.0} &
  \textbf{90.1} &
  \textbf{76.5} &
  \multicolumn{1}{c}{\cellcolor[HTML]{E3F2D9}\textbf{89.5}} &
  \textbf{91.7} &
  \textbf{92.5} &
  \textbf{84.8} &
  87.8 &
  95.0 \\
\rowcolor[HTML]{C9E4B4} 
\multirow{-6}{*}{\cellcolor[HTML]{F2F2F2}ViT-large} &
  \multirow{-3}{*}{\cellcolor[HTML]{F2F2F2}$384^2$} &
  Multi-label E2E &
  85.2 &
  89.8 &
  89.9 &
  76.4 &
  \multicolumn{1}{c}{\cellcolor[HTML]{C9E4B4}89.1} &
  91.6 &
  92.3 &
  84.0 &
  \textbf{89.5} &
  \textbf{96.1} \\ \hline

\end{tabular}
}
\end{table*}

%% file: tables/koleo.tex
\begin{table}[t]
\centering
\caption{Evaluation of $k$-NN-based entropy of penultimate-layer features (Euclidean distance). Models trained with our multi-label supervision (IN1k-Mul) consistently exhibit higher entropy across datasets, indicating greater feature diversity and reduced representation collapse compared to single-label training (IN1k-Sig). This suggests improved representational quality and helps explain the observed gains in transfer learning performance. 
}
\label{tab:koleo}
\resizebox{0.42\textwidth}{!}{
\begin{tabular}{c|l|ccc}
\hline
\multirow{2}{*}{\textbf{Model}} & \multicolumn{1}{c|}{\multirow{2}{*}{\textbf{Supervision}}} & \multicolumn{3}{c}{\textbf{Entropy $\uparrow$}}                     \\ \cline{3-5} 
                               & \multicolumn{1}{c|}{}                                      & \textbf{ImageNet} & \textbf{VOC} & \textbf{COCO} \\ \hline
\multirow{3}{*}{ResNet-50} & Rand. Init. & 1114          & 1540          & 1509          \\
                           & IN1k-Sig    & 1705          & 2582          & 2310          \\
                           & IN1k-Mul    & \textbf{2034} & \textbf{2663} & \textbf{2426} \\ \hline
\multirow{3}{*}{ViT-B}     & Rand. Init. & 0.0           & 137           & 19            \\
                           & IN1k-Sig    & 562           & 910           & 810           \\
                           & IN1k-Mul    & \textbf{791}  & \textbf{1040} & \textbf{959}  \\ \hline
\end{tabular}
}
\end{table}

%% file: tables/soft_label_training.tex
\begin{table}[t]
\caption{Top-1 accuracy comparison of soft-label training under different supervision strategies. We follow the ReLabel setup to generate a label map for each image and assign soft labels to random crops based on patchwise logits. All methods use ResNet-50 trained on ImageNet-1K with either 100 or 300 epochs. Our spatial label map achieves comparable or better classification accuracy, while requiring less labeled data.
}
\label{tab:soft_label}
\centering
\scriptsize
\resizebox{0.46\textwidth}{!}{
\begin{tabular}{l|cc|cc}
\hline
\multicolumn{1}{c|}{\multirow{2}{*}{\textbf{Method}}} & \multicolumn{2}{c|}{\textbf{100 Epoch}}   & \multicolumn{2}{c}{\textbf{300 Epoch}}    \\
\multicolumn{1}{c|}{}                                      & \textbf{IN-Val} & \textbf{ReaL} & \textbf{IN-Val} & \textbf{ReaL} \\ \hline
Original   & 77.5 & 83.5     & 78.2 & 84.1  \\
ReLabel~\cite{relabel}         & 71.9 & 78.6 & 78.9 & 85.0 \\
Classifier Map (Ours) & 74.9 & 82.1 & 78.9 & 84.9 \\ \hline
\end{tabular}
}
\end{table}

%% file: tables/filter_maskcut.tex
\begin{table}[t]
\caption{Zero-shot segmentation performance on COCO using CutLER masks filtered by our classification head. We remove pseudo-masks whose top-1 softmax confidence falls below threshold $\tau$. Despite excluding up to $12\%$ of masks (at $\tau=0.5$), segmentation performance improves across most metrics, suggesting that filtering out low-confidence proposals strengthens the quality of supervision for unsupervised segmentation. 
Metrics follow COCO-style evaluation: AP is the mean average precision over IoU thresholds, AP50/AP75 are AP at IoU 0.50/0.75, and APs, APm, APl represent AP for small, medium, and large objects, respectively. 
}
\label{tab:filter_maskcut}
\resizebox{0.48\textwidth}{!}{
\begin{tabular}{c|c|cccccc}
\hline
\textbf{Config} & \textbf{\# Masks} & \textbf{AP$\uparrow$}   & \textbf{AP50$\uparrow$}  & \textbf{AP75$\uparrow$} & \textbf{APs$\uparrow$}  & \textbf{APm$\uparrow$}  & \textbf{APl$\uparrow$} \\ \hline
Full & 1.93M (100\%)  & 8.71 & 17.50 & 7.78 & 1.91 & 7.33 & 22.52          \\
$\tau=0.3$  & 1.81M (94.0\%) & 8.81 & 17.80 & 7.90 & 1.87 & 7.49 & \textbf{22.52} \\
$\tau=0.5$  & 1.70M (88.0\%)    & \textbf{8.82} & \textbf{17.80} & \textbf{7.92} & \textbf{1.94} & \textbf{7.63} & 22.35        \\ \hline
\end{tabular}
}
\end{table}

%% file: figures/annotation_tool.tex
\begin{figure}[t]
    \centering
    \includegraphics[width=0.48\textwidth]{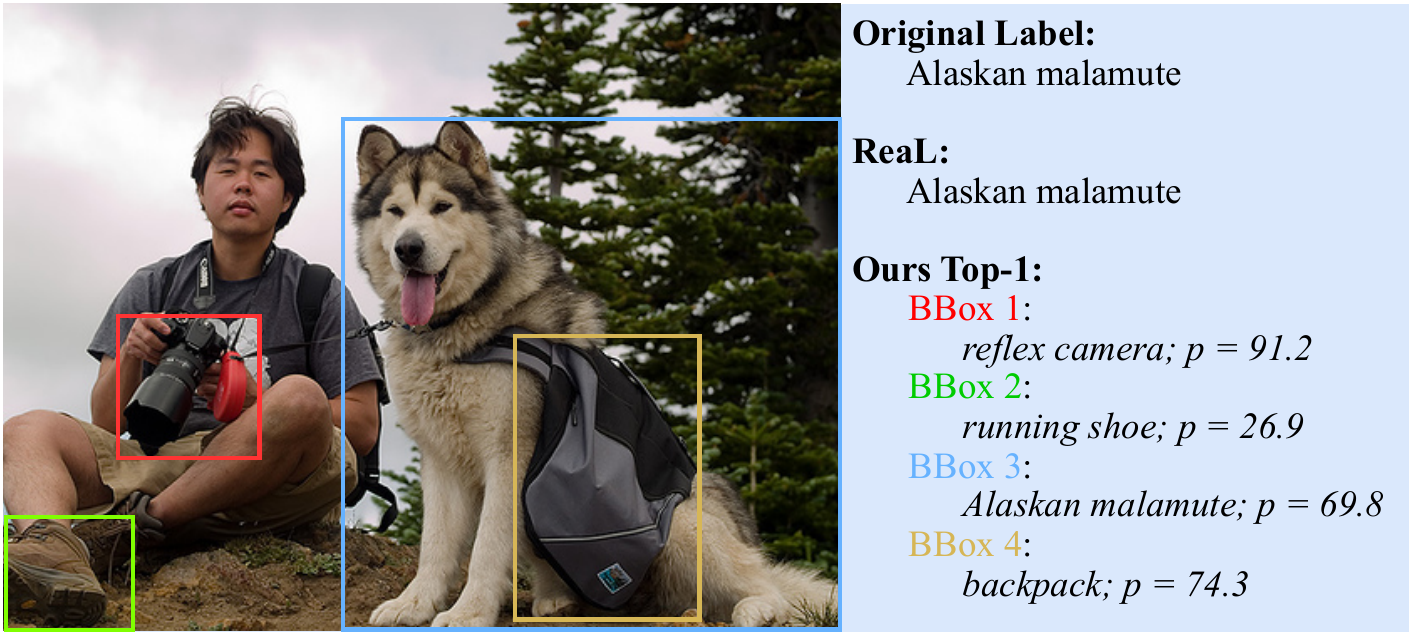}
    \caption{Example of interactive annotation using our classification head. Given arbitrary region proposals (highlighted in red), the model predicts ImageNet class labels with confidence scores. Even for out-of-vocabulary objects that are not explicitly covered by the label set (e.g., hiking boot), our classifier predicts semantically related categories (e.g., running shoe) with lower confidence, enabling efficient and flexible region-level labeling.}
    \label{fig:annotation_tool}
\end{figure}